%% file: main.tex
\documentclass[10pt,twocolumn,letterpaper]{article}

\usepackage[pagenumbers]{iccv} % To force page numbers, e.g. for an arXiv version

\input{preamble}

\definecolor{iccvblue}{rgb}{0.21,0.49,0.74}
\usepackage[pagebackref,breaklinks,colorlinks,allcolors=iccvblue]{hyperref}
\usepackage{balance}

\def\paperID{8929} % *** Enter the Paper ID here
\def\confName{ICCV}
\def\confYear{2025}

\title{egoPPG: Heart Rate Estimation from Eye-Tracking Cameras in \\ Egocentric Systems to Benefit Downstream Vision Tasks}

\author{Björn Braun,
Rayan Armani,
Manuel Meier,
Max Moebus,
and Christian Holz\\
Department of Computer Science, ETH Zürich, Switzerland\\
{\small\href{https://siplab.org/projects/egoPPG}{\color{magenta}{\texttt{https://siplab.org/projects/egoPPG}}}}%
{}
\vspace{-5mm}
}

\usepackage{xspace}
\def\projtask{\textit{egoPPG}\xspace}

\def\projmethod{\textit{PulseFormer}\xspace}
\def\projmethodnormal{PulseFormer\xspace}
\def\projdataset{\textit{egoPPG-DB}\xspace}
\newcommand{\projdatasetnormal}{egoPPG-DB\xspace}

\begin{document}

\twocolumn[{%
\renewcommand\twocolumn[1][]{#1}%
\maketitle
\begin{center}
    \centering
    \captionsetup{type=figure}
    \includegraphics[width=\textwidth]{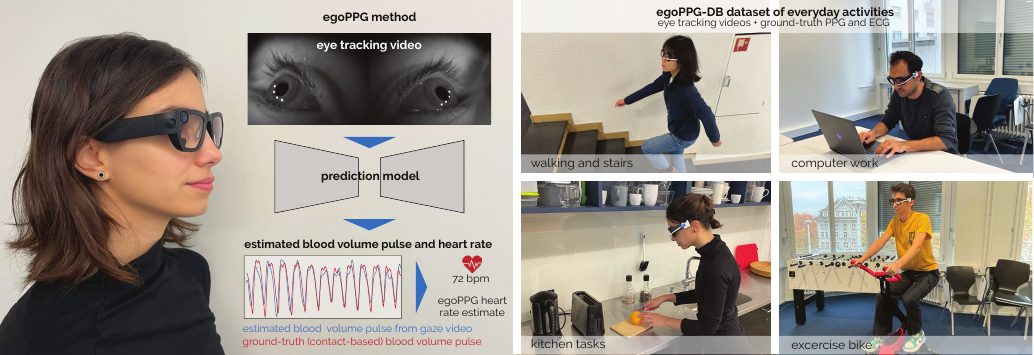}\vspace{-1mm}%
    \captionof{figure}{We propose \projtask as a novel computer vision task: tracking a person's heart rate (HR) on unmodified egocentric vision headsets.
    Taking eye tracking videos as input, our method \projmethod estimates the photoplethysmogram (PPG) from areas around the eyes to derive HR values.
    For training and validation, we introduce \projdataset, a dataset of eye tracking videos while participants performed everyday activities with synchronized ground-truth PPG (via nose-based contact sensor) and HR values (via ECG chest strap).\vspace{3mm}}
    \label{fig:teaser}
\end{center}%
}]

\input{sec/0_abstract}
\input{sec/1_introduction}
\input{sec/2_related_work}
\input{sec/3_overview}
\input{sec/4_dataset}
\input{sec/5_methods}
\input{sec/6_proficiency}
\input{sec/7_experiments}
\input{sec/8_discussion}
\input{sec/9_conclusion}

% Bib
{
    \small
    \balance
    \bibliographystyle{ieeenat_fullname}
    \bibliography{main}
}

% WARNING: do not forget to delete the supplementary pages from your submission 
\input{sec/X_suppl}

\end{document}

%% file: preamble.tex
\newcommand{\RNum}[1]{\uppercase\expandafter{\romannumeral #1\relax}}
\definecolor{darkgreen}{rgb}{0.167, 0.486, 0.347}
\definecolor{deemph}{gray}{0.55}
\usepackage{colortbl}
\newcommand{\grayrow}{\rowcolor[gray]{.9}}
\usepackage{pifont}% http://ctan.org/pkg/pifont
\usepackage{multirow, adjustbox}

%% file: sec/0_abstract.tex
\begin{abstract}
Egocentric vision systems aim to understand the spatial surroundings and the wearer's behavior inside it, including motions, activities, and interactions. 
We argue that egocentric systems must additionally detect physiological states to capture a person's attention and situational responses, which are critical for context-aware behavior modeling.
In this paper, we propose \emph{\projtask}, a novel vision task for egocentric systems to recover a person's cardiac activity to aid downstream vision tasks.
We introduce \emph{\projmethod}, a method to extract heart rate as a key indicator of physiological state from the eye tracking cameras on unmodified egocentric vision systems.
\projmethod continuously estimates the photoplethysmogram (PPG) from areas around the eyes and fuses motion cues from the headset's inertial measurement unit to track HR values.
We demonstrate \projtask's downstream benefit for a key task on EgoExo4D, an existing egocentric dataset for which we find \projmethod's estimates of HR to improve proficiency estimation by 14\%.
To train and validate \projmethod, we collected a dataset of 13+ hours of eye tracking videos from Project Aria and contact-based PPG signals as well as an electrocardiogram (ECG) for ground-truth HR values.
Similar to EgoExo4D, 25 participants performed diverse everyday activities such as office work, cooking, dancing, and exercising, which induced significant natural motion and HR variation (44--164\,bpm). 
Our model robustly estimates HR (MAE=7.67\,bpm) and captures patterns (r=0.85).
Our results show how egocentric systems may unify environmental and physiological tracking to better understand users and that \projtask as a complementary task provides meaningful augmentations for existing datasets and tasks.
We release our code, dataset, and HR augmentations for EgoExo4D to inspire research on physiology-aware egocentric tasks. 

\end{abstract}

%% file: sec/1_introduction.tex
\section{Introduction}
\label{sec:introduction}

% 1) Egocentric systems and tasks
Egocentric vision systems, such as Mixed Reality (MR) glasses by Meta~\cite{meta_quest}, Magic Leap~\cite{magic_leap}, and others % vendors~\cite{vivens}, 
have emerged as powerful devices for capturing and analyzing a person's behavior and their environment from a first-person perspective.
% Significant advances in vision-based perception algorithms for tasks like localization~\cite{sattler2011fast, kendall2015posenet, shavit2021learning}, mapping (SLAM)~\cite{davison2003real, hubner2020evaluation, rosinol2023nerf}, action recognition\cite{ma2016going, yan2022multiview, zhao2023learning, kondratyuk2021movinets, wu2022memvit}, and hand-object pose estimation~\cite{shan2020understanding, grauman2022ego4d, zhu2023egoobjects, dunnhofer2023visual} now make it possible to automatically digitize human motions and interactions within their environment. 
The wider availability of promising wearable capture platforms has sparked a large amount of research on egocentric vision tasks for environment understanding and navigation~\cite{engel2023project}, including localization~\cite{sattler2011fast, kendall2015posenet, shavit2021learning}, and simultaneous localization and mapping~\cite{davison2003real, hubner2020evaluation, rosinol2023nerf}.
Since egocentric systems simultaneously capture parts of the wearer's behavior in addition to their environment, prior work has investigated egocentric action recognition~\cite{ma2016going, yan2022multiview, zhao2023learning, wu2022memvit} and hand-object interaction~\cite{shan2020understanding, grauman2022ego4d, zhu2023egoobjects} to understand user behavior.
Several large-scale datasets now accelerate data-driven research in this domain with multimodal data for training and evaluation (e.g., Ego4D~\cite{grauman2022ego4d}, Nymeria~\cite{ma2024nymeria}, EgoExo4D~\cite{grauman2024ego}).

% 2) Attention and intention
% Beyond spatial awareness, egocentric systems aim to understand the user's behavior, their attention, and intent~\cite{yamada2012attention, admoni2016predicting, min2021integrating}. 
% For instance, anticipating the user's next action is crucial for applications in navigation, personalized feedback, and autonomous assistance~\cite{dixon2013surgeons, yao2019egocentric, wang2023holoassist}. 
% Objects of interest are commonly estimated from analyzed gaze patterns~\cite{fathi2012learning, li2013learning, huang2018predicting, lai2022eye} to support behavioral analysis and social understanding~\cite{fathi2012social, jiang2022egocentric, grauman2022ego4d, grauman2024ego}.
% 3) Holistic% holistically modeling a person's behavior and intent requires knowledge of their physiological state, which influences 
Most recently, Meta's Project Aria~2 introduced a contact-based heart rate (HR) sensor, with which egocentric systems can gauge the wearer's cognitive performance, attention, and situational responses~\cite{eysenck2007anxiety, coombes2009attentional, macpherson2009importance, tyng2017influences, shuggi2019motor}. 
% Affective computing has, furthermore, shown that a person's physiological state also gives insights into a person's affect~\cite{picard2000affective, calvo2015oxford, miranda2018amigos, subramanian2016ascertain, abadi2015decaf}.
% Key components of physiological state include cardiovascular indicators such as heart rate (HR) and electrodermal activity~\cite{miranda2018amigos, subramanian2016ascertain, abadi2015decaf, braun2024sympcam},
Numerous additional conditions manifest in a person's HR, such as emotions, stress and fatigue~\cite{picard2000affective, calvo2015oxford, miranda2018amigos, subramanian2016ascertain, abadi2015decaf}---capturing these dynamics can thus benefit models of human behavior to enable a richer understanding of user behavior. % actions and behavior for adaptive systems.

% In this paper, we introduce \textit{\projtask}, a novel task for egocentric vision systems to accurately extract a person's HR measurements from the system's built-in sensors, specifically the eye tracking cameras in unmodified headsets.
% We then propose \textit{\projmethod}, a novel method that implements \textit{\projtask} on Project Aria glasses to demonstrate the benefit of \textit{\projtask} for existing downstream vision tasks on large egocentric datasets.
In this paper, we introduce a method to make such HR estimates available to many \emph{existing} egocentric systems and \emph{already recorded} large datasets, such as EgoExo4D~\cite{grauman2024ego} or Nymeria~\cite{ma2024nymeria}.
Our method \projmethod accurately recovers a person's HR from the eye tracking videos in egocentric headsets.
\projmethod first estimates the person's photoplethysmogram (PPG) from the subtle fluctuations in skin intensity due to pulsatile artery expansion beneath the surface following a blood volume pulse (BVP), in particular deriving it from regions around the wearer's eye for robust tracking.
% Leveraging the infrared (IR) illuminant of the eye tracker, it works across light conditions~\cite{lv2024aria}.
Our spatial attention module ensures that PPG is estimated from robust regions around the eye, while our cross-attention fusion with the system's inertial measurement unit (IMU) learns a motion-informed temporal attention to optimally weight the eye tracking images for more accurate PPG estimates in scenarios with heavy motion.

We validate our method's efficacy on a novel dataset that we collected to capture some of the activities included in large-scale egocentric datasets alongside physiological reference recordings. 
Our dataset \projdataset contains 13~hours of recordings from 25~participants, who wore Project Aria glasses and performed six real-world tasks with varying motion and intensity, causing their HR values to reach levels between 44--164\,bpm.

\subsection*{Downstream benefits for egocentric vision tasks}
A key contribution of our paper is that we demonstrate that knowing a person's continuous HR values benefits egocentric vision tasks downstream.
We augment an existing architecture with \projmethod's HR estimates and show its impact on EgoExo4D's proficiency estimation benchmark, which improves accuracy on this task by 14.1\%. 

% The sensing configuration of our task can be considered a hybrid between typical contact-based BVP sensors (\eg, those in smartwatches~\cite{applewatch, googlepixel, mukhopadhyay2014wearable, dunn2018wearables, chow2020accuracy, moebus2024personalized}) and rPPG methods that aim to extract HR from a person's face using a camera~\cite{huelsbusch2002contactless, verkruysse2008remote, poh2010non}.
% While egocentric glasses are body-worn much like wrist watches, they couple more loosely to the body and are subject to considerable motion artifacts.
% Unlike contact sensors, eye trackers observe the wearer's eye regions from a short distance and capture eye motions and blinks, leading to ambiguity and noise for capturing fluctuations in skin intensity.
% Unlike rPPG configurations, egocentric capture systems move with the wearer's body and head, and eye trackers use controlled illumination.

% This setup, however, introduces novel challenges.
% As the eyes typically move strongly during everyday situations and are closed entirely while blinking, estimating the BVP from the full eye tracking images would introduce substantial motion artifacts.
% In contrast, the skin around the eyes typically experiences considerably less motion.
% Additionally, variations in how the glasses are worn result in different skin areas around the eyes being visible (see \cref{fig:eyes_spa}). 
% Finally, as the glasses sit on the user's head, natural head movements can introduce substantial motion artifacts.

\subsection*{Contributions}

We summarize our key contributions as follows:
\begin{enumerate}
    \item \textit{\projtask} as a novel task and \projmethod as an HR estimation method for egocentric systems that operates on eye tracking videos.
    Our method robustly predicts continuous HR across a series of activities and interactions (MAE=7.67\,bpm), with a 23.8\% lower error than current state-of-the-art rPPG models~\cite{chen2018deepphys, yu2019physnet,liu2020multi, yu2022physformer, yan2024physmamba}.

    \item \textit{\projdataset}, a dataset of eye tracking videos and synchronized BVP (contact-based) and ECG recordings (chest strap-based) to verify all physiological signals.
    We captured these across diverse everyday activities that were inspired by those included in existing large-scale egocentric datasets, such as EgoExo4D~\cite{grauman2022ego4d,ma2024nymeria,grauman2024ego}.
    
    \item a validation of \textit{\projtask's} downstream benefits for egocentric vision tasks.
    We demonstrate the implications of our method \projmethod on the proficiency estimation benchmark of the EgoExo4D dataset, which increases the accuracy by 14.1\% when augmenting EgoExo4D with our continuously predicted HR values.
\end{enumerate}

%% file: sec/2_related_work.tex
\section{Related work}
\label{sec:related_work}

\textbf{Egocentric vision.} In recent years, research in egocentric vision has surged, driven by advances in AR/VR glasses~\cite{apple, vive, magic_leap, meta_quest, microsoft_hololens, engel2023project}, which provide new ways for understanding user interaction from a first-person perspective.
Much of this work has focused on tasks such as action recognition \cite{ma2016going, yan2022multiview, zhao2023learning, kondratyuk2021movinets, wu2022memvit} and anticipation \cite{girdhar2021anticipative, damen2022rescaling, wu2022memvit}, full-body pose estimation \cite{shiratori2011motion, jiang2022avatarposer, jiang2023egoposer},  responding to user needs \cite{ryoo2015robot, rodin2021predicting, yao2019egocentric}, and social behavior analysis \cite{fathi2012social, jiang2022egocentric, grauman2022ego4d}.
Additionally, tracking vital signs in AR/VR settings and for affective computing applications~\cite{picard2000affective, calvo2015oxford, miranda2018amigos, subramanian2016ascertain, abadi2015decaf} has become an important tool for understanding users’ physiological states~\cite{marin2020emotion}, their behavior, attention, and intent~\cite{yamada2012attention, admoni2016predicting, min2021integrating}.

\noindent\textbf{Physiological measurements.} Wearable sensors have had a tremendous impact on health monitoring in recent years, enabling continuous measurement of key physiological metrics, such as heart rate (HR), oxygen saturation, and activity levels~\cite{mukhopadhyay2014wearable, dunn2018wearables, chow2020accuracy, moebus2024nightbeat, meier2024wildppg}.
HR, in particular, is a key measure for assessing an individual's health and performance~\cite{fox2007resting, evrengul2006relationship, shaffer2017overview, kleiger2005heart, luong2022characterizing}.
% While wearable sensors, e.g. wrist-worn smartwatches, provide accurate HR measurements in also challenging scenarios (\eg, exercising), they are intrusive and can cause discomfort~\cite{knight2006wearability}.
In parallel to wearable sensors such as smartwatches, recent research has extensively explored using cameras as an unobtrusive, non-contact alternative for measuring HR, generally called remote photoplethysmography (rPPG)~\cite{huelsbusch2002contactless, verkruysse2008remote, poh2010non}.
% rPPG detects the BVP via subtle color changes in the skin caused by the BVP.
rPPG measures HR based on the BVP via subtle color changes of the skin.
Generally, rPPG methods can be broadly divided into traditional signal processing techniques~\cite{huelsbusch2002contactless, poh2010non, braun2023sympathetic, de2013robust, wang2016algorithmic} and deep learning-based approaches~\cite{chen2018deepphys, yu2019physnet, liu2020multi, braun2024suboptimal, yan2024physmamba, speth2023non}.
So far, rPPG has been mostly applied to facial videos with the camera and user being stationary, such as while sitting in front of a laptop, as it requires a continuous video feed of the same skin region.
This limitation is shown in current rPPG datasets, which primarily capture individuals in seated positions with either a stationary camera directed at their face~\cite{stricker2014pure, heusch2017reproducible, bobbia2019unsupervised, sabour2021ubfc} or requiring users to hold a smartphone steadily in front of their face~\cite{tang2023mmpd}.
As a result, rPPG is not feasible to be deployed in more dynamic settings.

\noindent\textbf{Eye tracking cameras.} Eye tracking in egocentric vision systems is mostly done using inward-facing cameras directed at the eyes~\cite{adhanom2023eye}.
Even during motion, eye tracking in VR devices demonstrated accurate performance showcasing that the cameras remain almost stationary \textit{relative to} the user's eyes~\cite{clay2019eye}.
Furthermore, IR illumination makes them robust to lighting variations and low-light conditions~\cite{lv2024aria}.
To the best of our knowledge, videos from eye tracking cameras have not yet been explored for HR estimation. % using the BVP despite their promise to enable unobtrusive HR measurements during everyday life.

%% file: sec/3_overview.tex
\section{Overview}
\label{sec:overview}

Our aim is to enable egocentric vision systems i)~to model a person's physiological state via continuously estimated HR and ii)~to integrate these HR estimates into downstream tasks that benefit from knowledge of the user's state.
\cref{sec:dataset} first describes our dataset of synchronized eye tracking videos and ground-truth HR measurements. % during a series of everyday activities.
\cref{sec:methods} introduces our method \textit{\projmethod} for  recovering continuous HR from eye tracking videos.
% \cref{fig:architecture} illustrates our approach.
\cref{sec:downstream_proficiency} outlines the downstream benefits of our novel task, using HR as input for modeling user proficiency on the EgoExo4D dataset.
% Finally, \cref{sec:experiments} provides all results from our evaluations and \cref{sec:discussion} discusses our findings.

%% file: sec/4_dataset.tex
\section{\projdatasetnormal}
\label{sec:dataset}

The \textit{\projdataset} dataset was developed to support HR estimation from eye tracking videos under real-world conditions and contains significant motion and HR fluctuations. 
% The \textit{\projdataset} dataset was developed to support HR estimation from eye tracking videos under real-world conditions and with a protocol designed to elicit significant motion and fluctuations in HR.
By including diverse everyday activities, we provide a challenging benchmark for egocentric HR estimation models.

\subsection{Recruiting and recording}
We recruited $N=25$\,participants (12\,female, 13\,male, ages 19--32, $\mu=25.1$ and $\sigma=3.3$) on a voluntary basis, resulting in over 13\, hours of video recordings.
Based on the Fitzpatrick scale~\cite{fitzpatrick1988validity}, 9 participants had skin type \RNum{2}, 8 had skin type \RNum{3}, 3 had skin type \RNum{4}, and 5 had skin type \RNum{5}.
All participants signed a consent form before the data collection, agreeing with using and sharing their data for academic and non-commercial purposes.
% Participants were instructed to avoid wearing makeup prior to the recording. 
The data collection was approved by the ETH Zurich Ethics Commission (no. 2023-N-08).
In terms of duration, \textit{\projdataset} is among the longest rPPG datasets as listed in \cref{tab:appendix_rPPGdatasets} (Supplementary).
Participants of \textit{\projdataset} are not included in EgoExo4D.

\subsection{Apparatus}
\begin{figure}
    \centering
    \includegraphics[width=\columnwidth]{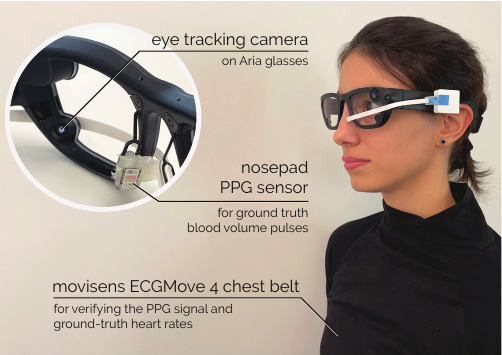}
    \caption{Apparatus used to record the \textit{\projdataset} dataset.}
    \label{fig:dataset_apparatus}
\end{figure}

\cref{fig:dataset_apparatus} illustrates our experimental setup. 
We used Project Aria glasses~\cite{engel2023project} with Profile 21 to record eye tracking videos at 30 fps with a resolution of $320 \times 240$ pixels per eye.
To capture ground truth PPG measurements, with which we train our model, we developed a custom sensor that records PPG data offline at 128\,Hz. 
The sensor consists of a main board, mounted on the left side of the frame, featuring a DA14695 system-on-chip interfacing with a MAX86141ENP+ PPG sensor. 
The LEDs and photodiodes used by the PPG sensor are embedded in the left nose pad and connected to the main board using a flat flexible cable. 
For each participant, we individually adjusted the nose pad position to ensure the sensor aligned with their left angular artery~\cite{holz2017glabella}. 
% By recording the PPG from the nose, instead of the wrist, we reduce the domain shift between the ground truth BVP (nose) and target BVP (skin region around the eyes).
To validate our custom PPG sensor, we also recorded gold-standard ECG data using a movisens ECGMove 4 chest belt sampling at 1024 Hz.
We synchronized all devices at the start and end of each recording with a synchronization pattern, using their built-in IMUs.
%, with sampling rates of 1000\,Hz, 128\,Hz and 64\,Hz for the Aria glasses, custom sensor, and ECGMove, respectively. 
% At the start and end of each recording, we placed the devices on top of each other and hit them against a flat surface three times to create a synchronization pattern.

\subsection{Capture protocol}
The average recording lasted 32\,minutes.
The capture protocol comprised 5 activities (\cref{tab:study_protocol}): watching a video, office work, kitchen work, dancing, and exercising on an indoor bike (\cref{fig:teaser}).
We included these activities for three purposes:
(1)~Incorporate everyday activities including the corresponding HR changes and motion artifacts;
(2)~cover a wide range of HR values (low HR when watching a video vs. high HR when exercising), and
(3)~resemble activities that were captured in large-scale egocentric vision datasets, such as EgoExo4D~\cite{grauman2024ego}.
In \cref{tab:appendix_detailed_tasks} (Supplementary), we give a detailed description of all activities. \cref{tab:results_activity} shows mean HR values for each activity.
Exercising on the bike produced the highest mean HR values (113\,bpm), whereas watching the video resulted in the lowest (71\,bpm).

\subsection{Dataset and signal quality verification}
To ensure the contact PPG sensor, whose signal we later use as the target for model training, produces accurate HR values, we evaluate it against the gold-standard ECG.
We calculated the MAE and Pearson correlation between HR estimates from the ECG and PPG signals for each participant using a 30-second sliding window.
For activity labeling, we manually annotated the start and end times of each task (see \cref{tab:study_protocol}) for each participant using the Point of View (POV) RGB videos recorded by the Project Aria glasses. 
To ensure that the signal quality of the contact PPG is sufficient for model training, we excluded all tasks with an MAE over 3.0\,bpm between the PPG and ECG (e.g. when the PPG sensor moved). %, which can happen when the PPG sensor occasionally loses alignment with the angular artery due to movement.
This applied to 20 of 150 tasks (13\%, see \cref{tab:appendix_excludedtasks} in Supplementary).
During the remaining tasks, our custom-built PPG nose sensor achieved very high accuracy, with an MAE of 1.3\,bpm and a correlation of 0.94 compared to the ECG signal, showing its suitability as ground truth.

\begin{table}[t]
    \centering
    \begin{tabular}{lll}
        \toprule[1.5pt]
        \textbf{Activity} & \textbf{Actions} & \textbf{Minutes} \\
        \midrule
        \midrule
        Watch video                     & Watch a documentary           & 5\\[2pt]
        \multirow{3}{*}{Office work}    & Work on a computer            & 4\\
                                        & Write on a paper              & 2\\
                                        & Talk to the experimenter      & 2\\[2pt]
        Walking                         & Walk to the kitchen           & 1\\[2pt]
        \multirow{3}{*}{Kitchen work}   & Cut vegetables                & \multirow{3}{*}{5}\\
                                        & Prepare a sandwich            & \\
                                        & Wash the dishes               & \\[2pt]
        Walking                         & Walk to the dancing room      & 1.5\\[2pt]
        Dancing                         & Follow random dance video     & 5\\[2pt]
        Exercise bike                   & Ride an exercise bike         & 5\\[2pt]
        Walking                         & Walk back to the start        & 1.5\\
        \bottomrule[1.5pt]
    \end{tabular}
    \caption{Capture protocol for recording the \textit{\projdataset} dataset.}
    \label{tab:study_protocol}
\end{table}

%% file: sec/5_methods.tex
\section{\projmethodnormal~method}
\label{sec:methods}
\label{sec:methods_hr}

\subsection{Problem definition}
Our objective is to estimate BVP and HR from periodic changes in pixel intensity in eye tracking video frames $\boldsymbol{F} \in \mathbb{R}^{w\times h}$. 
Physically, this means extracting physiological signals from the information in the light reflected by the arteries and arterioles that carry blood beneath the skin.
This light reflection can be modeled as a combination of diffuse and specular reflections. 
Wang \etal~\cite{wang2016algorithmic} model the reflected light intensity $C(t)$ as:
\begin{align}
    \boldsymbol{C(t)}=I(t) (\boldsymbol{v_s}(t)+\boldsymbol{v_d}(t))+\boldsymbol{v_n}(t)
\end{align}
where $I(t)$ is the luminance intensity, $\boldsymbol{v_s}(t)$ the specular reflection, $\boldsymbol{v_d}(t)$ the diffuse reflection, and $\boldsymbol{v_n(t)}$ the sensor noise.
While the specular reflection $\boldsymbol{v_s}(t)$ lacks pulsatile information, the diffuse reflection $\boldsymbol{v_d}(t)$ contains information about the absorption and scattering of the light in skin tissue~\cite{wang2016algorithmic}.
Thus, $\boldsymbol{v_d}(t)$ can be further decomposed as:
\begin{align}
     \boldsymbol{v_d}(t)=\boldsymbol{u_d} d_0+\boldsymbol{u_p} p(t)
\end{align}
where $\boldsymbol{u_d}$ is the unit color vector of the skin, $d_0$ the stationary reflection strength, $\boldsymbol{u_p}$ the relative absorption, and $p(t)$ the signals of interest.
$p(t)$ is in our case the BVP, which our model aims to learn from the camera recordings.

\begin{figure*}[t]
    \centering
    \includegraphics[width=\linewidth]{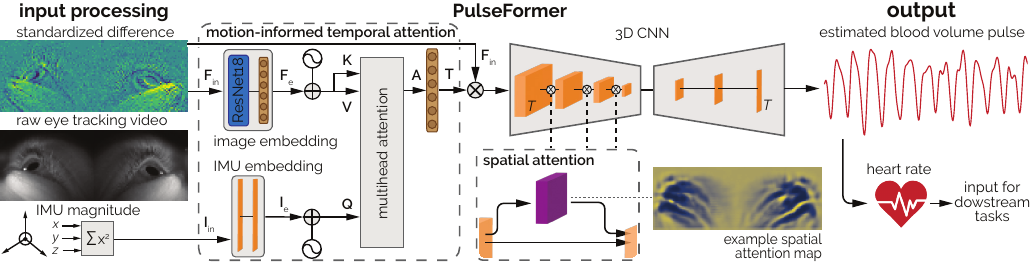}
    \caption{Architecture of our model for continuous BVP estimation from eye tracking videos and consecutive HR computation.}
    \label{fig:architecture}
\end{figure*}

\subsection{Deep learning model}
Our architecture is built upon a 3D CNN backbone (PhysNet)~\cite{yu2019physnet} with a temporal input length of $T=128$\,frames (corresponding to 4.3\,seconds) downsampled to $(h = 48) \times( w=128)$ pixels, resulting in an input of dimensions $(T,C,w,h)$.
The channel is $C = 1$ in our case, as our input is from monochrome videos.
The input in our network is the consecutive standardized frame differences (per participant frame-wise differences divided by the STD of the frames) of the eye tracking videos to help the network focus on the changes between frames~\cite{chen2018deepphys}.
% We standardize each frame by subtracting the mean pixel intensity and dividing it by the standard deviation of the pixel intensities values per participant~\cite{chen2018deepphys}.
As labels, we use the standardized consecutive differences of the PPG signals.
% Compared to the video of a person's face, which is usually used for rPPG tasks, eye tracking videos offer additional challenges.
Since eye tracking videos offer additional challenges compared to facial videos, usually used for rPPG tasks, we have designed our model to address these challenges (see \cref{fig:architecture}).

\noindent\textbf{Motion-informed temporal attention (MITA).} Egocentric glasses are body-worn and subject to considerable motion artifacts when the user moves.
Therefore, we propose to leverage the IMU within the glasses to obtain a motion-informed temporal attention. 
We employ a cross-attention module to integrate the IMU data with the video input, allowing our model to weigh each frame differently along the temporal dimension based on the motion intensity encoded by the IMU.
% An IMU measures the motion and orientation of the glasses, providing an indication of how much the user moves.
This allows the model, \eg, to give less emphasis to frames heavily affected by motion. 
Given the input feature map $\boldsymbol{F_{in}} \in \mathbb{R}^{T\times 1\times w\times h}$, we use ResNet18~\cite{he2016deep} and a linear layer to obtain the image embeddings $\boldsymbol{F_{e}} \in \mathbb{R}^{T \times D}$, where $D=128$ is the embedding dimension.
Given IMU measurements $\mathbf{I_{in}} \in \mathbb{R}^{T \times 1}$, we use two 1D convolutional layers to obtain the IMU embeddings $\mathbf{I_{e}} \in \mathbb{R}^{T \times D}$.
We then calculate the cross-attention $\mathbf{A} \in \mathbb{R}^{T \times D}$ as:
\begin{equation}
\mathbf{A} = \text{softmax}\left(\frac{\mathbf{Q}\mathbf{K}^\top}{\sqrt{D}}\right) \mathbf{V}
\end{equation}
where $\mathbf{I_{e}}$ serve as queries $Q$ and $\mathbf{F_{e}}$ as keys $K$ and values $V$.
Using a linear layer, we obtain the motion-informed temporal attention $\mathbf{T} \in \mathbb{R}^{T \times 1 \times 1 \times 1}$, which we multiply with $\boldsymbol{F_{in}}$.

\begin{figure}[t]
    \centering
    \includegraphics[width=\linewidth]{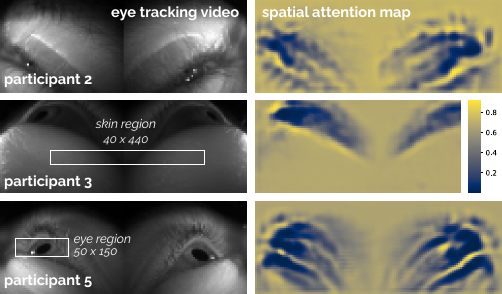}
    \caption{Left: Head geometry determines the regions that the eye tracker captures.
    Right: Learned spatial attention maps show that eye regions are excluded and \textit{\projmethod} instead extracts BVP from the surrounding skin regions, which moves less than the eyes.}
    \label{fig:eyes_spa}
\end{figure}

\noindent\textbf{Spatial attention (SA).} While the bulbar conjunctiva (white of the eyes) contains many blood vessels from which the BVP could theoretically be estimated, eyes typically move strongly during everyday situations and are closed while blinking (see participant 2 in \cref{fig:eyes_spa}).
Consequently, extracting the BVP from the eye regions would introduce substantial motion artifacts and reduce the signal-to-noise ratio (SNR).
In contrast, when qualitatively analyzing eye tracking images, we see that the skin around the eyes exhibits considerably less motion than the eyes themselves and could thus provide a more stable source of BVP information.
To address this, we introduce spatial attention modules~\cite{woo2018cbam, hu2021robust, niu2019robust} before each pooling (see~\cref{fig:architecture}) to allow our network to focus on high-SNR regions, such as the skin, and reduce the influence of low-SNR regions with frequent motion, like the eyes.
Given some feature map $\boldsymbol{F} \in \mathbb{R}^{T\times C\times w\times h}$, the spatial attention modules infer a spatial attention map $\boldsymbol{M_s} \in \mathbb{R}^{T\times 1\times w \times h}$ as:
\begin{align}
    \boldsymbol{M_s}(\boldsymbol{F}) = \sigma*(f^{7\times7}([\boldsymbol{F_{avg}};\boldsymbol{F_{max}}]))
\end{align}
where $\sigma$ is the sigmoid function, $f^{7\times7}$ a $7\times7$ convolution operation and  $\boldsymbol{F_{avg}} \in \mathbb{R}^{T\times 1\times w \times h}$ and $\boldsymbol{F_{max}}  \in \mathbb{R}^{T\times 1\times w \times h}$ are the average-pooled and max-pooled feature maps respectively.
% See \cref{fig:architecture} for an example learned spatial attention map.
The final output $\boldsymbol{F_{out}} \in \mathbb{R}^{T\times C\times w\times h}$ of each attention process is then the product of $\boldsymbol{M_s}$ and $\boldsymbol{F}$.
% \begin{equation}
% \mathbf{F_{out}} = \mathbf{M_s} \otimes \mathbf{F},
% \end{equation}

\noindent\textbf{Data augmentation.} Furthermore, individual variations in the fit of the glasses result in different parts of the skin around the eyes being visible.
For some individuals, the eye tracking cameras capture only the areas above the eyes, for others, only below, and in some cases, the glasses sit at an incline (see \cref{fig:eyes_spa}).
To account for such variations, we apply three targeted data augmentations during training that reflect these specific differences in camera angles and coverage: 
(1) random rotation between -20 and +20 degrees to account for slight inclinations; % in the glasses’ positioning; 
(2) random horizontal cropping to help the network distinguish between high and low SNR regions across various skin areas and camera positions; and 
(3) horizontal and vertical flipping to further increase robustness to differences in skin region visibility.
% Finally, to help the network focus on the changes between the frames caused by the periodic BVP, we use the standardized consecutive frame differences of the eye tracking videos as input into our network.
% We standardize each frame by subtracting the mean pixel intensity and dividing it by the standard deviation of the pixel intensities values~\cite{chen2018deepphys}.
% We use the standardized consecutive differences of the PPG signals from the nose as the labels for our model.
% \cref{fig:architecture} shows our architecture with the input preprocessing demonstrated on the left side, and an example learned spatial attention map on the right side.
Our model requires approximately 399\,GFLOPS per batch and has about 12M parameters.
The frame rate is 2.9k\,fps on an RTX 4090 and 180\,fps on an AMD EPYC CPU.

\subsection{Experiments setup}

% \subsubsection{Training}
\noindent\textbf{Training.}
We trained all models using five-fold cross-validation split by participants to ensure a strict separation between training, validation, and test sets.
We iteratively held out the data from five participants (20\%) as the test set, two as validation, and trained on the remaining with a batch size of 4 for 100 epochs, a learning rate of 0.0009, and mean squared error (MSE) as loss. 
In addition to our model, we used ten state-of-the-art rPPG baseline networks to compare the performance of our proposed model to these established models.
Our model was trained on a GeForce RTX 4090, with a total runtime of about 20\,hours for all folds.

\noindent\textbf{Evaluation.}
To calculate the HR, we filter the predicted BVP with a Butterworth filter (0.5–2.8\,Hz) and then detect peaks.
To assess model accuracy, we use the mean absolute error (MAE), root mean squared error (RMSE), mean absolute percentage error (MAPE), and Pearson correlation (r) using a non-overlapping 60-second window~\cite{de2013robust, liu2020multi, liu2024rppg}.
% We also provide insights into optimal hyperparameter configurations by analyzing the impact of image size, input length and preprocessing (see \cref{tab:appendix_hyperparameter}).
% Additionally, we calculate the Pearson correlation between predicted and ground truth values. Although correlation is not commonly used in rPPG, it provides valuable insight given the wide range of heart rates in our dataset, induced by activities like indoor cycling. This allows us to assess the model's ability to handle diverse HR variations.\\

\noindent\textbf{Video sampling rate.}
While we recorded the eye tracking videos with 30\,fps, large-scale datasets such as EgoExo4D~\cite{grauman2024ego} or Nymeria~\cite{ma2024nymeria} used only 10\,fps. 
To assess the impact of reduced fps, we evaluated model performance when (1) downsampling our videos to 10\,fps by retaining only every third frame and (2) downsampling to 10\,fps, then linearly interpolating between frames to upsample to 30\,fps.
% For both scenarios, we train from random initialization.

% \subsubsection{Hyperparameter optimization}
% We also provide insights into optimal hyperparameter configurations by analyzing the impact of image size, input length, loss function, and choice of label (see \cref{tab:appendix_hyperparameter}).

%% file: sec/6_proficiency.tex
\section{Downstream use for proficiency estimation}
\label{sec:downstream_proficiency}

To demonstrate the utility of predicting a user's physiological state for egocentric vision applications, we use the user proficiency estimation benchmark from the EgoExo4D dataset, which contains over 5000 videos from 740 participants performing skilled human activities~\cite{grauman2024ego}.
This benchmark aims to classify the proficiency of a user (novice, early expert, intermediate expert, late expert) using only egocentric videos (\textit{Ego}), only exocentric videos (\textit{Exo}), or all videos together (\textit{Ego + Exo}).
Our goal was to assess if we can improve the performance of the current baseline model (TimeSFormer~\cite{bertasius2021space}) when integrating our predicted HR data into the network.
This results in three additional configurations: using egocentric/exocentric videos and HR (\textit{Ego + HR}/\textit{Exo + HR}) and using all videos and HR (\textit{Ego + Exo + HR}).
To predict the continuous HRs for all EgoExo4D videos, we use \textit{\projmethod}, pre-trained on \textit{\projdataset}.

We implement the TimeSFormer model in exactly the same configuration as for the benchmark results~\cite{grauman2024ego} with a clip size of 16 frames and a sampling rate of 16, trained for 15 epochs. % epochs on four GeForce RTX 4090.
We use all videos of the EgoExo4D dataset, for which the proficiency estimation labels are available (using the official benchmark training/validation sets) and which have at least 16 frames at a sampling rate of 16, resulting in 2044 videos.
From the official training set, we use 10\% as validation, and the held-out official validation set for testing.
We summarize our predicted HR data by calculating five features (mean, STD, minimum and maximum HR, and mean HR change) for the corresponding videos.
% For each video, we calculate five features based on our predicted HR: mean, STD, minimum and maximum HR, and mean HR change.
We integrate these features via normalization and a 50-parameter linear layer whose output we concatenate with the output of TimeSFormer's backbone before feeding it into the classification head.
% To integrate our HR predictions into the TimeSFormer architecture, we simply feed our calculated features into a 50-parameter linear layer and concatenate the output with the backbone’s output before feeding it into the classification head.
We train all models from random initialization and evaluate using top-1 accuracy per EgoExo4D protocol.

%% file: sec/7_experiments.tex
\section{Experiments}
\label{sec:experiments}

\subsection{Heart rate estimation}
\label{sec:experiments_hr}

\subsubsection{Signal-processing baseline}
We employed signal processing to verify that the BVP signal is present in the eye tracking videos, to determine in which regions the SNR is highest, and to establish a baseline (see \cref{tab:results_hr}).
Since the glasses remain mostly stable throughout the recording, we manually define two spatial cropping regions per participant.
One region that includes mostly skin, and one region that includes mainly eyes (see \cref{fig:eyes_spa}).
We calculate the mean pixel intensity for both regions, remove motion artifacts by discarding any changes outside the interquartile range and finally filter the signal with a 4\textsuperscript{th} order Butterworth bandpass filter (0.5 to 2.8\,Hz) to obtain the BVP (see \cref{fig:appendix_raw_signals} in Supplementary).

\subsubsection{\projmethodnormal method} Using our proposed network \projmethod, we obtain an MAE of 7.67\,bpm and a correlation of 0.85 between our predicted HR and the ground truth HR (see \cref{tab:results_hr}).
This is an improvement of 2.40\,bpm (23.8\%) of the MAE and 0.13 for the correlation compared to the current SOTA FactorizePhys~\cite{joshi2024factorizephys}.
Split by activity, we obtain the lowest MAE while the participants are watching a video (MAE=5.52\,bpm) and the highest MAE during exercising on a bike (MAE=12.91\,bpm), which is the task with the highest mean HR (113.1\,bpm) and the second highest motion magnitude.
In addition, the MAE decreases for all activities when adding MITA, with the greatest performance improvement for dancing.
We define the motion magnitude as the root-mean-squared sum of the absolute differences across the 3-axis IMU recorded by the Aria glasses and normalize it between zero and one across all activities to get a measure of motion of each activity.
See \cref{fig:appendix_boxplot} (Supplementary) for a boxplot of the MAEs of \projmethod's predictions.
Using signal processing, we obtain an MAE of 12.40\,bpm when using the skin region around the eyes and an MAE of 14.60\,bpm using the eye regions as input.
This is also reflected in the spatial attention maps that our model implicitly learns (see \cref{fig:eyes_spa}), which exclude the eyes to predict the HR.
To qualitatively cross-check these results, \cref{fig:appendix_raw_signals} (Supplementary) shows an example plot of the raw mean intensity values (before filtering) of the skin region compared to the eye region, with the BVP clearly visible for the skin region.
\cref{tab:results_hr_fps} shows the results when downsampling our videos to 10\,fps.
MAE increases to 11.13\,bpm and the correlation decreases to 0.7 when training and testing using 10\,fps.
When upsampling the videos again to 30\,fps, the MAE decreases to 10.20\,bpm and the correlation increases to 0.77.
In \cref{sec:supp_cross_dataset} (Supplementary), we show that \projmethod also outperforms the baselines in a cross-dataset evaluation.

\addtolength{\tabcolsep}{-0.2em}
\begin{table}
    \centering
        \begin{tabular}{@{}lrccc@{}}
            \toprule[1.5pt]
            \textbf{Model} & \textbf{MAE} & \textbf{RMSE} & \textbf{MAPE} & \textbf{r} \\
            \midrule
            \midrule
            Yue et al.~\cite{yue2023facial}             & 29.63 & 32.99 & 37.86 & 0.1 \\
            DeepPhys~\cite{chen2018deepphys}            & 28.26 & 31.97 & 36.68 & 0.08 \\
            TS-CAN~\cite{liu2020multi}                  & 26.32 & 32.39 & 29.13 & 0.11 \\
            ContrastPhys+~\cite{sun2024contrast}        & 19.12 & 24.13 & 22.57 & 0.21 \\
            RhythmMamba~\cite{zou2024rhythmmamba}       & 15.05 & 19.78 & 17.46 & -0.16 \\
            Baseline eyes                               & 14.60 & 18.18 & 18.37 & 0.20 \\
            PhysMamba~\cite{yan2024physmamba}           & 13.94 & 16.86 & 17.76 & 0.61 \\
            RhythmFormer~\cite{zou2024rhythmformer}     & 13.13 & 17.43 & 14.73 & 0.51 \\
            Baseline skin                               & 12.40 & 15.54 & 15.29 & 0.50 \\
            PhysNet~\cite{yu2019physnet}                & 12.09 & 15.43 & 15.14 & 0.66 \\
            PhysFormer~\cite{yu2022physformer}          & 10.71 & 13.97 & 12.69 & 0.72 \\
            \grayrow \projmethod w/o SA        & 10.49 & 13.62 & 12.83 & 0.73 \\
            FactorizePhys~\cite{joshi2024factorizephys} & 10.07 & 13.43 & 12.36 & 0.67 \\
            \grayrow \projmethod w/o MITA      & 8.82  & 12.03 & 10.82 & 0.81 \\
            \grayrow \textbf{\projmethod (ours)}& \textbf{7.67} & \textbf{10.69} & \textbf{9.45} & \textbf{0.85} \\
            \midrule
            Improvement over & \multirow{2}{*}{\textcolor{darkgreen}{\texttt{-}\textbf{2.40}}} & \multirow{2}{*}{\textcolor{darkgreen}{\texttt{-}\textbf{2.74}}} & \multirow{2}{*}{\textcolor{darkgreen}{\texttt{-}\textbf{2.91}}} & \multirow{2}{*}{\textcolor{darkgreen}{\texttt{+}\textbf{0.13}}} \\
            % Improvement & \textcolor{darkgreen}{\texttt{-}\textbf{3.0}} & \textcolor{darkgreen}{\texttt{-}\textbf{3.3}} & \textcolor{darkgreen}{\texttt{-}\textbf{3.2}} & \textcolor{darkgreen}{\texttt{+}\textbf{0.13}} \\
            second-best method &&&&\\
            \bottomrule[1.5pt]
        \end{tabular}
    \caption{Results for HR prediction from eye tracking videos using different models (\projmethod, \projmethod without SA, \projmethod without MITA and established rPPG baselines).}
    \label{tab:results_hr}
\end{table}
\addtolength{\tabcolsep}{0.2em}

\addtolength{\tabcolsep}{-0.4em}
\begin{table}
    \centering
    \begin{tabular}{@{}lrccc@{}}
        \toprule[1.5pt]
        % \textbf{Activity} & \textbf{$\mu$ HR} & \textbf{Motion} & \textbf{MAE} & \textbf{RMSE} & \textbf{MAPE} \\
        \multirow{2}{*}{\textbf{Activity}} & \multirow{2}{*}{\textbf{$\mu$ HR}} & \textbf{Motion} & \multirow{2}{*}{\textbf{\projmethod}} & \textbf{\projmethod} \\
        && \textbf{magnitude} && \textbf{w/o MITA}\\
        \midrule
        \midrule
        Video       & 71.5             & 0             & \textbf{5.52}  & \textbf{5.97}   \\
        Office      & 75.7             & 0.45          & 7.50   & 8.22 \\
        Kitchen     & 85.3             & 0.54          & 7.22   & 8.89  \\
        Dancing     & 89.1             & \textbf{1.00} & 7.85   & 10.54  \\
        Bike        & \textbf{113.1}   & 0.77          & 12.91  & 14.62  \\
        Walking     & 93.7             & 0.30          & 8.23   & 8.29  \\
        \bottomrule[1.5pt]
    \end{tabular}
    \caption{Results for HR prediction (MAE) split by activity using \projmethod and \projmethod without MITA. }
    \label{tab:results_activity}
\end{table}
\addtolength{\tabcolsep}{0.4em}

% \addtolength{\tabcolsep}{-0.4em}
% \begin{table}
%     \centering
%     \begin{tabular}{@{}lrcrrr@{}}
%         \toprule[1.5pt]
%         % \textbf{Activity} & \textbf{$\mu$ HR} & \textbf{Motion} & \textbf{MAE} & \textbf{RMSE} & \textbf{MAPE} \\
%         \multirow{2}{*}{\textbf{Activity}} & \multirow{2}{*}{\textbf{$\mu$ HR}} & \textbf{Motion} & \multirow{2}{*}{\textbf{MAE}} & \multirow{2}{*}{\textbf{RMSE}} & \multirow{2}{*}{\textbf{MAPE}} \\
%         && \textbf{magnitude} &&&\\
%         \midrule
%         \midrule
%         Video       & 71.5             & 0             & \textbf{5.5$\pm$1.6}  & \textbf{8.2$\pm$3.1}   & 8.9$\pm$3.5 \\
%         Office      & 75.7             & 0.45          & 7.5$\pm$1.1   & 10.4$\pm$1.6  & 10.9$\pm$2.0 \\
%         Kitchen     & 85.3             & 0.54          & 7.2$\pm$0.9   & 9.6$\pm$2.1   & 8.8$\pm$1.5 \\
%         Dancing     & 89.1             & \textbf{1.00} & 7.9$\pm$2.0   & 9.9$\pm$2.3   & 8.7$\pm$2.3 \\
%         Bike        & \textbf{113.1}   & 0.77          & 12.9$\pm$2.7  & 16.3$\pm$2.9  & 10.5$\pm$2.0 \\
%         Walking     & 93.7             & 0.30          & 8.2$\pm$3.3   & 10.2$\pm$3.9  & \textbf{7.9$\pm$2.8} \\
%         \bottomrule[1.5pt]
%     \end{tabular}
%     \caption{Results for HR prediction split by activity. }
%     \label{tab:results_activity}
% \end{table}
% \addtolength{\tabcolsep}{0.4em}

\addtolength{\tabcolsep}{-0.1em}
\begin{table}
    \centering
    \begin{tabular}{@{}lcccc@{}}
        \toprule[1.5pt]
        \textbf{Input video} & \textbf{MAE} & \textbf{RMSE} & \textbf{MAPE} & \textbf{r} \\
        \midrule
        \midrule
        10\,fps (other datasets)    & 11.13 & 15.18 & 12.28 & 0.70 \\
        Upsampled to 30\,fps        & \textbf{10.18} & \textbf{13.07} & \textbf{12.48} & \textbf{0.77} \\
        \bottomrule[1.5pt]
    \end{tabular}
    \caption{   Results for HR prediction with different frame rates. 
                In the first row, we downsample our videos to a frame rate of 10\,fps, commonly used by large-scale datasets such as EgoExo4D~\cite{grauman2024ego}.
                In the second row, we first downsample our videos to 10\,fps and then upsample them to 30\,fps by linearly interpolating between frames.}
    \label{tab:results_hr_fps}
\end{table}
\addtolength{\tabcolsep}{0.1em}

\subsection{Downstream task: proficiency estimation}
\label{sec:experiments_proficiency}
\cref{tab:results_timesformer} summarizes the results of our experiments to evaluate the value of HR estimation for the proficiency estimation benchmark on EgoExo4D.
We see that integrating our predicted HRs into the TimeSFormer model~\cite{bertasius2021space} improved accuracy for all scenarios but one (Soccer).
We also achieved the highest accuracy for each of these individual scenarios with our HR integration.
When combining the egocentric videos with our predicted HRs, we achieved an overall accuracy of 45.29\%, a 14.1\% increase compared to using egocentric videos alone. 
The largest gains appeared in the cooking and dancing tasks, where accuracy rose from 20.00\% to 40.00\% and from 43.44\% to 53.27\%. 
Also, when using the egocentric and exocentric videos, and our predicted HRs together, the accuracy increased by 12.67\% (4.94 percentage points) from 39.00\% to 43.94\% compared to using only the egocentric and exocentric videos.
% Using only exocentric videos yielded the lowest overall accuracy at 35.9\%.

\begin{table*}
    \centering
    \begin{tabular}{@{}lccccccc@{}}
        \toprule[1.5pt]
        \textbf{Scenario} & \textbf{Majority} & \textbf{Ego} & \textbf{Ego + HR (ours)} & \textbf{Exo} & \textbf{Exo + HR (ours)} & \textbf{Ego + Exo} & \textbf{Ego + Exo + HR (ours)}\\
        % &&& \textbf{(ours)} & & \textbf{(ours)} && \textbf{HR (ours)} \\
        \midrule
        \midrule
        Basketball  & 38.00 & 45.45 & 47.47             & 48.48             & 48.48         & 49.49 & \textbf{50.50}     \\
        Cooking     & 0.00  & 20.00 & \textbf{40.00}    & 35.00             & \textbf{40.00}& 25.00 & \textbf{40.00}     \\ 
        Dancing     & 24.59 & 43.44 & 53.27             & 42.62             & 48.36         & 50.82 & \textbf{59.84}     \\
        Music       & 57.89 & 78.94 & \textbf{81.58}    & 57.89             & 57.89         & 57.89 & 60.53     \\
        Bouldering  & 15.29 & 24.50 & \textbf{27.81}    & 8.61              & 12.58         & 15.89 & 21.19     \\
        Soccer      & 62.50 & 50.00 & 56.25             & \textbf{81.25}    & 75.00         & 75.00 & 62.50     \\
        \midrule
        Overall     & 27.80 & 39.69 & \textbf{45.29}    & 34.75             & 37.67         & 39.00 & 43.94     \\
        \bottomrule[1.5pt]
    \end{tabular}
    \caption{Results for proficiency estimation benchmark on EgoExo4D dataset.
    Note that for all scenarios except Soccer, the accuracy increases when integrating \projmethod's heart rate estimate into the existing and otherwise unmodified baseline model.}
    \label{tab:results_timesformer}
\end{table*}

%% file: sec/8_discussion.tex
\section{Discussion}
\label{sec:discussion}

\subsection{Heart rate estimation}
\label{sec:discussion_hr}
Evaluating \textit{\projmethod} on \textit{\projdataset}, we showed that HR can be reliably predicted from eye tracking videos and IMU signals from unmodified egocentric vision headsets.
% Compared to the performances achieved on popular rPPG datasets such as PURE~\cite{stricker2014pure}, UBFC-RPPG~\cite{bobbia2019unsupervised}, and MMPD~\cite{tang2023mmpd}, we achieved very competitive performance in challenging settings~\cite{liu2024rppg}. 
% While the lowest reported MAE for UBFC-RPPG in the rPPG-toolbox~\cite{liu2024rppg} is 1.2\,bpm, its participants sit almost motion-free with their eyes closed.
% Already on MMPD, a dataset with varied lighting and little motion, the lowest reported MAE increases to 10.2\,bpm~\cite{liu2024rppg}, as it incorporates more realistic challenges due to recordings on mobile devices with light motion (head rotation, talking, and taking selfies).
While SOTA rPPG models (e.g., PhysFormer) achieve MAEs as low as 0.50\,bpm~\cite{yu2022physformer, zou2024rhythmformer} on datasets such as UBFC-RPPG~\cite{bobbia2019unsupervised} or OBF~\cite{li2018obf}, these datasets captured participants while \emph{calmly sitting} at a table looking at the camera (with very little motion or HR changes).
However, during even just \emph{light} motion (e.g., on MMPD~\cite{tang2023mmpd} or VIPL-HR~\cite{niu2019vipl}), their MAE increases to 5.0--12.0\,bpm~\cite{yu2022physformer, zou2024rhythmformer}.

On \textit{\projdataset}, which contains much stronger motion (dancing, exercise bike) and HR fluctuations (between 44--164\,bpm), \textit{\projmethod's} MAE is 7.67\,bpm and outperforms the rPPG SOTA FactorizePhys (MAE=10.07\,bpm).
Given the strong motion and diverse everyday activities in \textit{\projdataset}, we believe that our results demonstrate the robustness of \textit{\projmethod} in dynamic, everyday conditions.
For context, even HR estimates from contact sensors tightened to the body (e.g., Apple Watch) yield an MAE of 3.0\,bpm during rest and an MAE of 4.6\,bpm on a bike~\cite{gillinov2017variable}.

\subsubsection{Performance depending on method}
We introduced MITA and leverage SA modules to improve the performance of our model.
The performance of \textit{\projmethod} decreases from 7.67\,bpm to 8.82\,bpm when removing the MITA and to 10.49\,bpm when removing the SA modules (see \cref{tab:results_hr}).
When qualitatively analyzing the learned SA maps, we see that our model implicitly learned to exclude the eyes for estimating BVP from the eye tracking videos (see \cref{fig:eyes_spa}).
This aligns with our results using signal processing, obtaining better performance for the skin region compared to the eyes (see \cref{tab:results_hr}).

\subsubsection{Performance depending on activity}
Analyzing our results split by activity (see \cref{tab:results_activity}), we obtain the highest MAE when exercising on a bike (MAE=12.91\,bpm) and the lowest MAE when watching a video (MAE=5.52\,bpm). 
While watching a video yields the lowest MAE, it is higher than MAEs typically reported for rPPG datasets, such as UBFC-RPPG~\cite{bobbia2019unsupervised}, despite similar levels of motion. 
We attribute this to two factors: first, the higher variability in HR across \projdataset, requiring improved generalization, and second, the inherent motion artifacts in eye tracking videos from blinking and natural eye movements, even during static tasks like watching videos.
Such inherent motion artifacts and, e.g., slipping glasses can make capturing rPPG more difficult in this manner.
Furthermore, although dancing has the highest motion magnitude, its MAE (7.85\,bpm) is comparable to that of lower-motion tasks such as office and kitchen activities.
When comparing performance with and without our MITA module, we observe an improvement of 2.6\,bpm for dancing, indicating that MITA effectively addresses motion-induced artifacts.

\subsubsection{Performance depending on camera fps}
Using eye tracking videos recorded at only 10\,fps considerably decreases performance (see \cref{tab:results_hr_fps}).
However, upsampling the frame rate to 30\,fps through linear interpolation between frames substantially improves the performance again.
This is especially important as many large-scale datasets, such as EgoExo4D~\cite{grauman2024ego} or Nymeria~\cite{ma2024nymeria}, for which predicting a user's physiological state could help for further downstream tasks, are recorded at only 10\,fps.

\subsection{Benefits for proficiency estimation downstream}
\label{sec:discussion_proficiency}
We found that incorporating HR data into the baseline model of the proficiency estimation task substantially improved accuracy across all three configurations. 
The egocentric videos combined with the HR achieved the highest overall accuracy at 45.29\%, marking a 14.1\% increase over using only egocentric videos (39.69\%).
Adding HR especially improved accuracy for cooking (from 20\% to 40\%) and dancing (from 43.44\% to 53.27\%), which had the lowest accuracies besides bouldering when using only egocentric videos, demonstrating the value of HR in enhancing model performance.
Combining egocentric videos, exocentric videos, and HR provided further accuracy gains for some scenarios, achieving the best results for basketball, cooking, and dancing.
Results using exocentric views alone were lower overall, which is consistent with benchmark results~\cite{grauman2024ego}. 
Soccer was the only scenario, for which the performance decreased for \textit{Exo+HR} and \textit{Ego+Exo+HR}.
We see two reasons for that.
1)~Of EgoExo4D's 2044 official train/test videos, only 77 are soccer, making it the scenario with the least training/test data by far.
2)~Our HR estimates may be less accurate for ``Stop-and-Go'' sports, which are not captured in \textit{\projdataset} right now.
% Our findings suggest that adding HR data as an auxiliary signal enhances performance for the proficiency estimation benchmark on the EgoExo4D dataset.
In \cref{tab:downstream_egoexo4d_comparison} (Supplementary), we show that we obtain the best downstream performance when using the HR features calculated with \projmethod compared to the baselines.
For training and testing, we used the available subset of EgoExo4D videos for which proficiency labels are available, following the official training and test splits. 
While our used data shows slight variations from the official release in majority class distributions and accuracy scores, the observed trends align well with the established benchmark results.

% \subsection{Limitations and future work}
% \label{sec:discussion_limitations_future}
% % We observed the highest MAE during biking, having the highest HRs.
% % We believe that promising approaches to address these limitations could be to record more tasks with high HRs.
% % Future studies could extend data collection to outdoor settings to assess the impact of varying lighting conditions.
% While we ensured a balanced gender ratio (13 male, 12 female), our sample size of 25 participants restricts broader demographic conclusions. 
% In future work, we aim to expand the dataset to investigate performance variations across different age groups, skin types, and ethnicities.
% Furthermore, future studies could extend data collection to outdoor settings to assess the impact of varying lighting conditions and record more tasks with higher HRs.
% Finally, we believe that it is a highly interesting problem to explore further downstream applications of a user's physiological state, such as personalized feedback, autonomous assistance, as well as health-related applications. 

\subsection{Broader impacts}
\label{sec:broader_impacts}
Beyond health applications, such as predicting stress and fatigue~\cite{picard2000affective, moebus2024meaningful, calvo2015oxford}, cardiac measurements could also help models better understand user behavior to, \eg, improve personalized assistance~\cite{kovacevic2024multimodal}.
Furthermore, we believe that our approach requires the user's explicit consent, regardless of application.
Mechanisms must make users aware of measurements and require consent, \eg, on the Aria platform.

%% file: sec/9_conclusion.tex
\section{Conclusion}
\label{sec:conclusion}
\textit{\projtask} is a novel task for egocentric vision systems to extract the wearer's heart rate for integrating their physiological state into egocentric vision tasks downstream. % by accurately estimates a person's HR from eye tracking cameras. 
We have introduced \textit{\projmethod}, a method that processes input from the eye tracking cameras on unmodified egocentric vision systems and fuses them with motion cues from the headset’s IMU to robustly estimate the person's HR in various everyday scenarios. 
We validate \textit{\projmethod's} robustness on our dataset \textit{\projdataset} and demonstrate significant improvements over existing rPPG models.
With HR estimations from \textit{\projmethod}, we also significantly improve the proficiency estimation benchmark on the large-scale EgoExo4D dataset.
% With \textit{\projmethod}, we can estimate HR on the large-scale EgoExo4D dataset and show that integrating HR estimations into the proficiency estimation benchmark of EgoExo4D increases prediction accuracy by 14.1\%.
Our results emphasize the potential of physiological insights obtained via \textit{\projtask} methods for further egocentric vision applications.
By making our dataset available to the community, we aim to support physiological state estimation via HR in future research and new downstream tasks for egocentric vision systems.
% While we ensured a balanced gender ratio (13 male, 12 female), our sample size of 25 participants restricts broader demographic conclusions. 
Given our promising results, we believe that future work could now focus on collecting more participants with a broader demographic background across different age groups, skin types, and ethnicities, and also extend data collection to outdoor settings to assess the impact of varying lighting conditions.

%% file: sec/X_suppl.tex
\clearpage
\setcounter{page}{1}
\maketitlesupplementary

%%%%%%%%%%%%%%%%%%%%%%%%%%%%%%%%%%%%%%%%%%%%%%%%%%%%%%%%%%%%%%%%%%%%%%%%%%%%%%%%%%%%%%%%%%%%%%%%%%
\section{Related datasets}
\cref{tab:appendix_rPPGdatasets} gives a comparison of the dataset size and activities of some related remote photoplethysmography (rPPG) datasets.
In terms of hours of recordings and recorded frames, \textit{\projdataset} is among the largest dataset.
Furthermore, we see that all comparable rPPG datasets only include activities with very little motion and heart rate (HR) changes such as watching videos, head rotations or talking.
In contrast, \textit{\projdataset} features a wide variety of challenging everyday activities, such as kitchen work, dancing and riding an exercise bike, which induce significant motion artifacts and HR changes.

%%%%%%%%%%%%%%%%%%%%%%%%%%%%%%%%%%%%%%%%%%%%%%%%%%%%%%%%%%%%%%%%%%%%%%%%%%%%%%%%%%%%%%%%%%%%%%%%%%
\section{Excluded tasks}
For all participants and activities, we checked the mean absolute error (MAE) between the predicted HR from our custom contact PPG sensor on the nose and the gold standard ECG from the chest belt.
We excluded all tasks with an MAE over 3.0 beats per minute (bpm), which can happen, for example, when the PPG sensor loses alignment with the angular artery due to movement.
In this way, we ensured that the photoplethysmography (PPG) signal from the nose, which we used as the target signal to train our model, is highly accurate.
As a result, we had to exclude 20 out of the 150 tasks (13\%), which we list in \cref{tab:appendix_excludedtasks}.
We can see that this applied only to tasks with more motion (dancing, exercise bike, and walking).
Since the participants had to walk multiple stairs throughout the data recording, this mostly happened during walking.

\begin{table}[ht]
    \centering
    \begin{tabular}{@{}ll@{}}
        \toprule[1.5pt]
        \textbf{Activity} & \textbf{Excluded participants} \\
        \midrule
        \midrule
        Watch video  & ---\\
        Office work  & ---\\
        Kitchen work  & ---\\
        Dancing  & 012, 015 \\
        Exercise bike  & 009, 012, 014, 015, 016, 023 \\
        Walking  & 004, 012, 013, 014, 018, 021, 022\\
        \bottomrule[1.5pt]
    \end{tabular}
    \caption{Detailed table of all excluded tasks.}
    \label{tab:appendix_excludedtasks}
\end{table}

%%%%%%%%%%%%%%%%%%%%%%%%%%%%%%%%%%%%%%%%%%%%%%%%%%%%%%%%%%%%%%%%%%%%%%%%%%%%%%%%%%%%%%%%%%%%%%%%%%
\section{Detailed description of activities}
\cref{tab:appendix_detailed_tasks} gives a comprehensive description of the actions for each activity during our recording.
Generally, participants were free to talk during the entire duration of the recording and conduct the tasks as they would do it normally.
For example, during the kitchen work, the participants were completely free to prepare the sandwich and if they would like to eat or drink while doing it.

%%%%%%%%%%%%%%%%%%%%%%%%%%%%%%%%%%%%%%%%%%%%%%%%%%%%%%%%%%%%%%%%%%%%%%%%%%%%%%%%%%%%%%%%%%%%%%%%%%
\section{Data recording}
\begin{figure*}
    \centering
    \includegraphics[width=\linewidth]{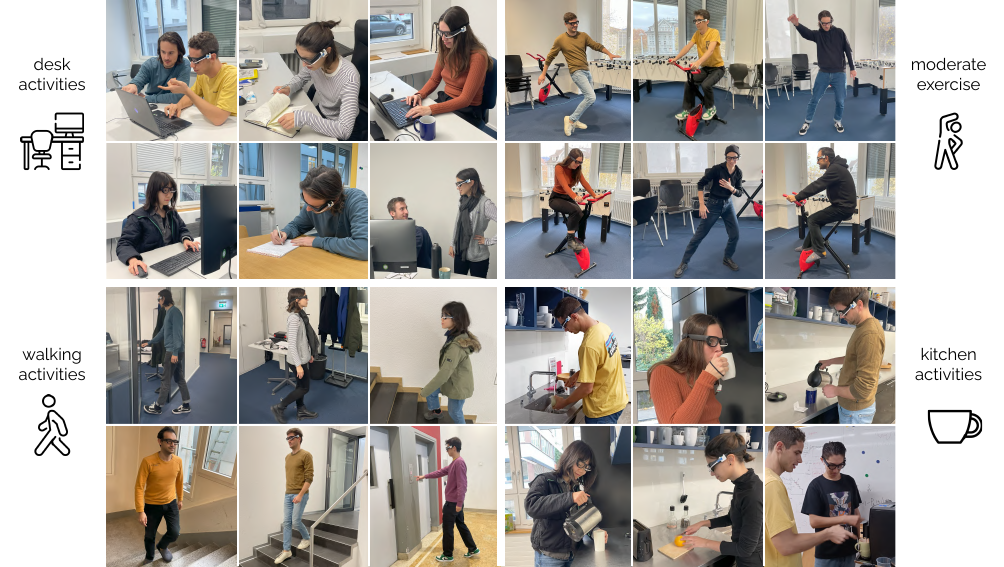}
    \caption{Additional images of the data recording showing the variety of everyday activities our dataset includes.}
    \label{fig:appendix_studyapparatus}
\end{figure*}

In \cref{fig:appendix_studyapparatus}, we show a variety of different images and people of our data recording from a third person view to visualize the apparatus and capture protocol.
All participants visible in these images explicitly agreed to be visualized.

%%%%%%%%%%%%%%%%%%%%%%%%%%%%%%%%%%%%%%%%%%%%%%%%%%%%%%%%%%%%%%%%%%%%%%%%%%%%%%%%%%%%%%%%%%%%%%%%%%
\section{Initial signal verification}
\begin{figure}[ht]
    \centering
    \includegraphics[width=\columnwidth]{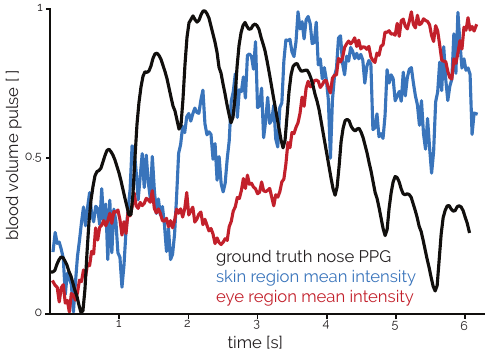}
    \caption{Example raw mean intensity of the skin and eye region, showing the higher SNR for the skin region around the eyes compared to the eyes.}
    \label{fig:appendix_raw_signals}
\end{figure}

In \cref{fig:appendix_raw_signals}, we show the raw mean intensity values after spatial cropping of the skin region and the eye region (see \cref{fig:eyes_spa}) compared to the ground truth contact PPG signal from the nose.
We can clearly see that the blood volume pulse is present both in the eyes and skin region with the skin region having a higher signal-to-noise ratio (SNR) compared to the eyes.

\section{Variance of results}
\begin{figure}[ht]
    \centering
    \includegraphics[width=\columnwidth]{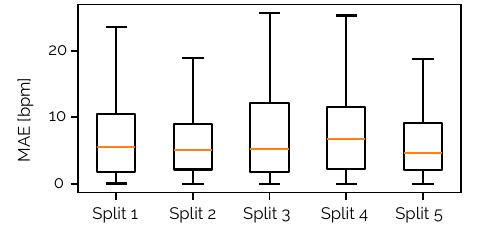}
    \caption{Boxplot of the MAEs of the predictions of \textit{\projmethod}.}
    \label{fig:appendix_boxplot}
\end{figure}

In \cref{fig:appendix_boxplot} we show the boxplot of the MAEs of the predictions of \textit{\projmethod} on \textit{\projdataset} by split.
The interquartile range across all splits is between 1.7 and 10.5\,bpm.

\section{Cross-dataset evaluation}
\label{sec:supp_cross_dataset}
We evaluated \projmethod and the two strongest baselines when training on three conventional rPPG datasets (MMPD~\cite{tang2023mmpd}, UBFC-rPPG~\cite{bobbia2019unsupervised}, and PURE~\cite{stricker2014pure}) and testing on \projdataset (\cref{tab:cross_dataset_on_egoppg}), and vice versa (\cref{tab:cross_dataset_on_conventional}).
For the rPPG datasets, we extracted the eye region using MediaPipe~\cite{lugaresi2019mediapipe}, resized to $48 \times 128$, and converted to grayscale. 
\projmethod consistently outperforms the baselines across all scenarios and datasets (except one case), showing strong generalization to unseen data.
Please note that we can only evaluate \projmethod w/o MITA as conventional rPPG datasets do not contain IMU data from the participants' heads.

% Cross dataset evaluation when training on other conventional rPPG datasets and testing on egoPPG-DB
\begin{table}[h]
    \centering
    \begin{tabular}{@{}llcc@{}}
        \textbf{Train Set} & \textbf{Model} & MAE & MAPE \\
        \midrule
                                    & PhysFormer & 20.56 & 27.06 \\
         MMPD                       & FactorizePhys & \multicolumn{2}{c}{Not converging} \\
                                    & \textbf{\projmethod w/o MITA} & \textbf{13.66} & \textbf{16.64} \\[3pt]

                                    & PhysFormer & 18.32 & 23.63 \\
         UBFC-rPPG                  & FactorizePhys & 18.58 & 24.46 \\
                                    & \textbf{\projmethod w/o MITA} & \textbf{14.83} & \textbf{18.57} \\[3pt]

                                    & PhysFormer & 24.39 & 24.94 \\
         PURE                       & FactorizePhys & 13.20 & 15.44 \\
                                    & \textbf{\projmethod w/o MITA} & \textbf{12.99} & \textbf{13.46} \\
    \end{tabular}
    \caption{Results (MAE) when training on conventional rPPG datasets and testing on \projdataset.}
    \label{tab:cross_dataset_on_egoppg}
\end{table}

% Cross dataset evaluation when training on egoPPG-DB and evaluating on other conventional rPPG datasets
\addtolength{\tabcolsep}{-0.4em}
\begin{table}[h]
    \centering
    \begin{tabular}{@{}lcccccc@{}}
        \multirow{2}{*}{\textbf{Model}} & \multicolumn{2}{c}{\textbf{MMPD}} & \multicolumn{2}{c}{\textbf{UBFC-rPPG}} & \multicolumn{2}{c}{\textbf{PURE}} \\
        & MAE & MAPE & MAE & MAPE & MAE & MAPE \\
        \midrule
        PhysFormer & 11.76 & 14.57 & 16.80 & 16.46 & 23.89 & 37.50 \\
        FactorizePhys & 12.06 & 15.11 & \textbf{14.28} & \textbf{14.98} & 26.10 & 40.62 \\
        \grayrow \textbf{\projmethod (ours)} & \textbf{11.48} & \textbf{15.08} & 15.09 & 15.81 & \textbf{23.56} & \textbf{36.71} \\
    \end{tabular}
    \caption{Results (MAE) when training on \projdataset and testing on conventional rPPG datasets.}
    \label{tab:cross_dataset_on_conventional}
\end{table}
\addtolength{\tabcolsep}{0.4em}

\section{HR distribution} 
\label{sec:suppl_hr_distributions}
\begin{figure}[h]
    \centering
    \includegraphics[width=1.0\linewidth]{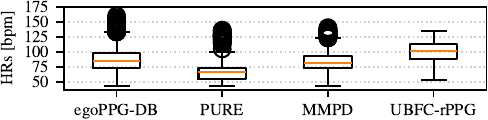}
    \caption{Boxplot of HRs of \projdataset and three rPPG datasets.}%
    \label{fig:boxplot_all_dataset}
\end{figure}

\projdataset exhibits the widest HR range (44–164\,bpm, see \cref{fig:boxplot_all_dataset}) and significantly more motion (e.g., dancing, exercise bike) than other evaluated rPPG datasets, where participants typically sit calmly at a table.

\section{Downstream performance comparison}
\label{sec:supp_downstream_comparison}
HR features from the other evaluated baselines perform progressively worse than those from \projmethod when used for proficiency estimation on EgoExo4D, highlighting the importance of accurate HR estimation for downstream tasks (see \cref{tab:downstream_egoexo4d_comparison}).

\begin{table}[h]
    \centering
    \begin{tabular}{@{}lccc@{}}
        \textbf{Model} & \textbf{Ego+HR} & \textbf{Exo+HR} & \textbf{Ego+Exo+HR} \\
        \midrule
         FactorizePhys      & 44.62 & 36.72 & 40.13 \\
         PhysFormer         & 44.39 & 36.66 & 43.07 \\
         \grayrow \textbf{\projmethod (ours)} & \textbf{45.29} & \textbf{37.67} & \textbf{43.94} \\
    \end{tabular}
    \caption{Downstream performance (accuracy) on EgoExo4D using the HR predictions from the three best baseline models.}
    \label{tab:downstream_egoexo4d_comparison}
\end{table}

%%%%%%%%%%%%%%%%%%%%%%%%%%%%%%%%%%%%%%%%%%%%%%%%%%%%%%%%%%%%%%%%%%%%%%%%%%%%%%%%%%%%%%%%%%%%%%%%%%
% \section{Hyperparameter optimization}
% In \cref{tab:appendix_hyperparameter}, we show the results of \textit{\projmethod} when training our model with different hyperparameters such as different image sizes, input window sizes, or target signals.

\addtolength{\tabcolsep}{-0.4em}
\begin{table*}
    \centering
    \begin{tabular}{@{}lllll@{}}
        \toprule[1.5pt]
        \textbf{Dataset} & \textbf{Part.} & \textbf{Frames} & \textbf{Hours} & \textbf{Tasks}\\
        \midrule
        \midrule
        PURE~\cite{stricker2014pure}                & 10    & 110\,K    &   1   & Resting, talking, small head movements\\
        MAHNOB-HCI~\cite{soleymani2011multimodal}   & 27    & 2.6\,M    &   12  & Watching videos\\
        MMPD~\cite{tang2023mmpd}                    & 33    & 1.2\,M    &   11  & Resting, head rotation, selfie videos\\
        MMSE-HR~\cite{zhang2016multimodal}          & 40    & 310\,K    &   2   & Talking, watching videos, experiencing different emotions\\
        UBFC-rPPG~\cite{bobbia2019unsupervised}     & 43    & 150\,K    &   1.5 & Gaming on a computer\\
        UBFC-PHYS~\cite{sabour2021ubfc}             & 56    & 2.4\,M    &   19  & Resting, Trier Social Stress Test\\
        OBF~\cite{li2018obf}                        & 106   & 3.8\,M    &   18  & Resting with varying HR levels \\
        VIPL-HR~\cite{niu2019vipl}                  & 107   & \textbf{4.3\,M}   &   \textbf{20}  & Resting, talking, head rotation, different lighting conditions\\
        SCAMPS (synthetic)~\cite{mcduff2022scamps}  & \textbf{2800} & 1.7\,M & 16 & Different facial actions \\
        &&&&\\
        \textit{\projdataset} (ours)                   & 25    & 1.4\,M    &   13  & Watching videos, office and kitchen work, dancing, biking, walking\\
        \bottomrule[1.5pt]
    \end{tabular}
    \caption{Summary of existing datasets for rPPG.}
    \label{tab:appendix_rPPGdatasets}
\end{table*}
\addtolength{\tabcolsep}{0.2em}

\begin{table*}
    \centering
    \begin{tabular}{lll}
        \toprule[1.5pt]
        \textbf{Activity} & \textbf{Actions} & \textbf{Description} \\
        \midrule
        \midrule
        Watch video                     & Watch a documentary           & Watch a relaxing documentary on a computer.\\
                                        &&\\
        \multirow{3}{*}{Office work}    & Work on a computer            & Randomly browse through websites and type text from a PDF into Word.\\
                                        & Write on a paper              & Write a text from a PDF on a computer onto a piece of paper.\\
                                        & Talk to the experimenter      & Have a free, unscripted conversation with the experimenter.\\
                                        &&\\
        Walking                         & Walk to the kitchen           & Walk along a hallway, down the stairs into the kitchen.\\
                                        &&\\
        \multirow{5}{*}{Kitchen work}   & Get ingredients               & Get all ingredients for a sandwich from the fridge.\\
                                        & Cut vegetables                & Get a cutting board, knife and a plate and cut vegetables.\\
                                        & Prepare a sandwich            & Put the bread into the toaster and afterward freely prepare sandwich.\\
                                        & Eat sandwich/drink            & Participants are free to eat the sandwich or drink during the recording.\\
                                        & Wash the dishes               & Wash everything used while preparing the sandwich. \\
                                        &&\\
        Walking                         & Walk to the dancing room      & Walking along a hallway into a new room for dancing and biking.\\
                                        &&\\
        Dancing                         & Follow random dance video     & Choose a dance video and afterward follow it.\\
                                        &&\\
        Exercise bike                   & Ride an exercise bike         & Ride an exercise bike with moderate to high intensity.\\
                                        &&\\
        \multirow{2}{*}{Walking}        & Walk back to the physical     & Walk back to the physical location of the start either up the stairs\\
        &location of the start & or using the elevator.\\
        \bottomrule[1.5pt]
    \end{tabular}
    \caption{Detailed capture protocol and action descriptions of the \textit{\projdataset} dataset.}
    \label{tab:appendix_detailed_tasks}
\end{table*}

%% file: main.bbl
\begin{thebibliography}{99}
\providecommand{\natexlab}[1]{#1}
\providecommand{\url}[1]{\texttt{#1}}
\expandafter\ifx\csname urlstyle\endcsname\relax
  \providecommand{\doi}[1]{doi: #1}\else
  \providecommand{\doi}{doi: \begingroup \urlstyle{rm}\Url}\fi

\bibitem[Abadi et~al.(2015)Abadi, Subramanian, Kia, Avesani, Patras, and Sebe]{abadi2015decaf}
Mojtaba~Khomami Abadi, Ramanathan Subramanian, Seyed~Mostafa Kia, Paolo Avesani, Ioannis Patras, and Nicu Sebe.
\newblock Decaf: Meg-based multimodal database for decoding affective physiological responses.
\newblock \emph{IEEE Transactions on Affective Computing}, 6\penalty0 (3):\penalty0 209--222, 2015.

\bibitem[Adhanom et~al.(2023)Adhanom, MacNeilage, and Folmer]{adhanom2023eye}
Isayas~Berhe Adhanom, Paul MacNeilage, and Eelke Folmer.
\newblock Eye tracking in virtual reality: a broad review of applications and challenges.
\newblock \emph{Virtual Reality}, 27\penalty0 (2):\penalty0 1481--1505, 2023.

\bibitem[Admoni and Srinivasa(2016)]{admoni2016predicting}
Henny Admoni and Siddhartha Srinivasa.
\newblock Predicting user intent through eye gaze for shared autonomy.
\newblock In \emph{2016 AAAI fall symposium series}, 2016.

\bibitem[Apple(2024)]{apple}
Apple.
\newblock Apple vision pro.
\newblock \url{https://www.apple.com/apple-vision-pro/}, 2024.
\newblock Accessed: 2024.11.13.

\bibitem[Bertasius et~al.(2021)Bertasius, Wang, and Torresani]{bertasius2021space}
Gedas Bertasius, Heng Wang, and Lorenzo Torresani.
\newblock Is space-time attention all you need for video understanding?
\newblock In \emph{ICML}, page~4, 2021.

\bibitem[Bobbia et~al.(2019)Bobbia, Macwan, Benezeth, Mansouri, and Dubois]{bobbia2019unsupervised}
Serge Bobbia, Richard Macwan, Yannick Benezeth, Alamin Mansouri, and Julien Dubois.
\newblock Unsupervised skin tissue segmentation for remote photoplethysmography.
\newblock \emph{Pattern Recognition Letters}, 124:\penalty0 82--90, 2019.

\bibitem[Braun et~al.(2023)Braun, McDuff, Baltrusaitis, and Holz]{braun2023sympathetic}
Bj{\"o}rn Braun, Daniel McDuff, Tadas Baltrusaitis, and Christian Holz.
\newblock Video-based sympathetic arousal assessment via peripheral blood flow estimation.
\newblock \emph{Biomedical Optics Express}, 14\penalty0 (12):\penalty0 6607--6628, 2023.

\bibitem[Braun et~al.(2024)Braun, McDuff, and Holz]{braun2024suboptimal}
Bj{\"o}rn Braun, Daniel McDuff, and Christian Holz.
\newblock How suboptimal is training rppg models with videos and targets from different body sites?
\newblock In \emph{Proceedings of the IEEE/CVF Conference on Computer Vision and Pattern Recognition}, pages 410--418, 2024.

\bibitem[Calvo et~al.(2015)Calvo, D'Mello, Gratch, and Kappas]{calvo2015oxford}
Rafael~A Calvo, Sidney D'Mello, Jonathan~Matthew Gratch, and Arvid Kappas.
\newblock \emph{The Oxford handbook of affective computing}.
\newblock Oxford University Press, USA, 2015.

\bibitem[Chen and McDuff(2018)]{chen2018deepphys}
Weixuan Chen and Daniel McDuff.
\newblock Deepphys: Video-based physiological measurement using convolutional attention networks.
\newblock In \emph{Proceedings of the european conference on computer vision (ECCV)}, pages 349--365, 2018.

\bibitem[Chow et~al.(2020)Chow, Yang, et~al.]{chow2020accuracy}
Hsueh-Wen Chow, Chao-Ching Yang, et~al.
\newblock Accuracy of optical heart rate sensing technology in wearable fitness trackers for young and older adults: validation and comparison study.
\newblock \emph{JMIR mHealth and uHealth}, 8\penalty0 (4):\penalty0 e14707, 2020.

\bibitem[Clay et~al.(2019)Clay, K{\"o}nig, and Koenig]{clay2019eye}
Viviane Clay, Peter K{\"o}nig, and Sabine Koenig.
\newblock Eye tracking in virtual reality.
\newblock \emph{Journal of eye movement research}, 12\penalty0 (1), 2019.

\bibitem[Coombes et~al.(2009)Coombes, Higgins, Gamble, Cauraugh, and Janelle]{coombes2009attentional}
Stephen~A Coombes, Torrie Higgins, Kelly~M Gamble, James~H Cauraugh, and Christopher~M Janelle.
\newblock Attentional control theory: Anxiety, emotion, and motor planning.
\newblock \emph{Journal of anxiety disorders}, 23\penalty0 (8):\penalty0 1072--1079, 2009.

\bibitem[Damen et~al.(2022)Damen, Doughty, Farinella, Furnari, Kazakos, Ma, Moltisanti, Munro, Perrett, Price, et~al.]{damen2022rescaling}
Dima Damen, Hazel Doughty, Giovanni~Maria Farinella, Antonino Furnari, Evangelos Kazakos, Jian Ma, Davide Moltisanti, Jonathan Munro, Toby Perrett, Will Price, et~al.
\newblock Rescaling egocentric vision: Collection, pipeline and challenges for epic-kitchens-100.
\newblock \emph{International Journal of Computer Vision}, pages 1--23, 2022.

\bibitem[Davison(2003)]{davison2003real}
Davison.
\newblock Real-time simultaneous localisation and mapping with a single camera.
\newblock In \emph{Proceedings Ninth IEEE International Conference on Computer Vision}, pages 1403--1410. IEEE, 2003.

\bibitem[De~Haan and Jeanne(2013)]{de2013robust}
Gerard De~Haan and Vincent Jeanne.
\newblock Robust pulse rate from chrominance-based rppg.
\newblock \emph{IEEE transactions on biomedical engineering}, 60\penalty0 (10):\penalty0 2878--2886, 2013.

\bibitem[Dunn et~al.(2018)Dunn, Runge, and Snyder]{dunn2018wearables}
Jessilyn Dunn, Ryan Runge, and Michael Snyder.
\newblock Wearables and the medical revolution.
\newblock \emph{Personalized medicine}, 15\penalty0 (5):\penalty0 429--448, 2018.

\bibitem[Engel et~al.(2023)Engel, Somasundaram, Goesele, Sun, Gamino, Turner, Talattof, Yuan, Souti, Meredith, et~al.]{engel2023project}
Jakob Engel, Kiran Somasundaram, Michael Goesele, Albert Sun, Alexander Gamino, Andrew Turner, Arjang Talattof, Arnie Yuan, Bilal Souti, Brighid Meredith, et~al.
\newblock Project aria: A new tool for egocentric multi-modal ai research.
\newblock \emph{arXiv preprint arXiv:2308.13561}, 2023.

\bibitem[Evrengul et~al.(2006)Evrengul, Tanriverdi, Kose, Amasyali, Kilic, Celik, and Turhan]{evrengul2006relationship}
Harun Evrengul, Halil Tanriverdi, Sedat Kose, Basri Amasyali, Ayhan Kilic, Turgay Celik, and Hasan Turhan.
\newblock The relationship between heart rate recovery and heart rate variability in coronary artery disease.
\newblock \emph{Annals of Noninvasive Electrocardiology}, 11\penalty0 (2):\penalty0 154--162, 2006.

\bibitem[Eysenck et~al.(2007)Eysenck, Derakshan, Santos, and Calvo]{eysenck2007anxiety}
Michael~W Eysenck, Nazanin Derakshan, Rita Santos, and Manuel~G Calvo.
\newblock Anxiety and cognitive performance: attentional control theory.
\newblock \emph{Emotion}, 7\penalty0 (2):\penalty0 336, 2007.

\bibitem[Fathi et~al.(2012)Fathi, Hodgins, and Rehg]{fathi2012social}
Alircza Fathi, Jessica~K Hodgins, and James~M Rehg.
\newblock Social interactions: A first-person perspective.
\newblock In \emph{2012 IEEE Conference on Computer Vision and Pattern Recognition}, pages 1226--1233. IEEE, 2012.

\bibitem[Fitzpatrick(1988)]{fitzpatrick1988validity}
Thomas~B Fitzpatrick.
\newblock The validity and practicality of sun-reactive skin types i through vi.
\newblock \emph{Archives of dermatology}, 124\penalty0 (6):\penalty0 869--871, 1988.

\bibitem[Fox et~al.(2007)Fox, Borer, Camm, Danchin, Ferrari, Lopez~Sendon, Steg, Tardif, Tavazzi, Tendera, et~al.]{fox2007resting}
Kim Fox, Jeffrey~S Borer, A~John Camm, Nicolas Danchin, Roberto Ferrari, Jose~L Lopez~Sendon, Philippe~Gabriel Steg, Jean-Claude Tardif, Luigi Tavazzi, Michal Tendera, et~al.
\newblock Resting heart rate in cardiovascular disease.
\newblock \emph{Journal of the American College of Cardiology}, 50\penalty0 (9):\penalty0 823--830, 2007.

\bibitem[Gillinov et~al.(2017)Gillinov, Etiwy, Wang, Blackburn, Phelan, Gillinov, Houghtaling, Javadikasgari, and Desai]{gillinov2017variable}
Stephen Gillinov, Muhammad Etiwy, Robert Wang, Gordon Blackburn, Dermot Phelan, A~Marc Gillinov, Penny Houghtaling, Hoda Javadikasgari, and Milind~Y Desai.
\newblock Variable accuracy of wearable heart rate monitors during aerobic exercise.
\newblock \emph{Medicine \& Science in Sports \& Exercise}, 2017.

\bibitem[Girdhar and Grauman(2021)]{girdhar2021anticipative}
Rohit Girdhar and Kristen Grauman.
\newblock Anticipative video transformer.
\newblock In \emph{Proceedings of the IEEE/CVF international conference on computer vision}, pages 13505--13515, 2021.

\bibitem[Grauman et~al.(2022)Grauman, Westbury, Byrne, Chavis, Furnari, Girdhar, Hamburger, Jiang, Liu, Liu, et~al.]{grauman2022ego4d}
Kristen Grauman, Andrew Westbury, Eugene Byrne, Zachary Chavis, Antonino Furnari, Rohit Girdhar, Jackson Hamburger, Hao Jiang, Miao Liu, Xingyu Liu, et~al.
\newblock Ego4d: Around the world in 3,000 hours of egocentric video.
\newblock In \emph{Proceedings of the IEEE/CVF Conference on Computer Vision and Pattern Recognition}, pages 18995--19012, 2022.

\bibitem[Grauman et~al.(2024)Grauman, Westbury, Torresani, Kitani, Malik, Afouras, Ashutosh, Baiyya, Bansal, Boote, et~al.]{grauman2024ego}
Kristen Grauman, Andrew Westbury, Lorenzo Torresani, Kris Kitani, Jitendra Malik, Triantafyllos Afouras, Kumar Ashutosh, Vijay Baiyya, Siddhant Bansal, Bikram Boote, et~al.
\newblock Ego-exo4d: Understanding skilled human activity from first-and third-person perspectives.
\newblock In \emph{Proceedings of the IEEE/CVF Conference on Computer Vision and Pattern Recognition}, pages 19383--19400, 2024.

\bibitem[He et~al.(2016)He, Zhang, Ren, and Sun]{he2016deep}
Kaiming He, Xiangyu Zhang, Shaoqing Ren, and Jian Sun.
\newblock Deep residual learning for image recognition.
\newblock In \emph{Proceedings of the IEEE conference on computer vision and pattern recognition}, pages 770--778, 2016.

\bibitem[Heusch et~al.(2017)Heusch, Anjos, and Marcel]{heusch2017reproducible}
Guillaume Heusch, Andr{\'e} Anjos, and S{\'e}bastien Marcel.
\newblock A reproducible study on remote heart rate measurement.
\newblock \emph{arXiv preprint arXiv:1709.00962}, 2017.

\bibitem[Holz and Wang(2017)]{holz2017glabella}
Christian Holz and Edward~J Wang.
\newblock Glabella: Continuously sensing blood pressure behavior using an unobtrusive wearable device.
\newblock \emph{Proceedings of the ACM on Interactive, Mobile, Wearable and Ubiquitous Technologies}, 1\penalty0 (3):\penalty0 1--23, 2017.

\bibitem[HTC(2024)]{vive}
HTC.
\newblock Htc vive.
\newblock \url{https://vive.com/}, 2024.
\newblock Accessed: 2024.11.13.

\bibitem[Hu et~al.(2021)Hu, Qian, Wang, He, Guo, and Ren]{hu2021robust}
Min Hu, Fei Qian, Xiaohua Wang, Lei He, Dong Guo, and Fuji Ren.
\newblock Robust heart rate estimation with spatial--temporal attention network from facial videos.
\newblock \emph{IEEE Transactions on Cognitive and Developmental Systems}, 14\penalty0 (2):\penalty0 639--647, 2021.

\bibitem[H{\"u}bner et~al.(2020)H{\"u}bner, Clintworth, Liu, Weinmann, and Wursthorn]{hubner2020evaluation}
Patrick H{\"u}bner, Kate Clintworth, Qingyi Liu, Martin Weinmann, and Sven Wursthorn.
\newblock Evaluation of hololens tracking and depth sensing for indoor mapping applications.
\newblock \emph{Sensors}, 20\penalty0 (4):\penalty0 1021, 2020.

\bibitem[Huelsbusch and Blazek(2002)]{huelsbusch2002contactless}
Markus Huelsbusch and Vladimir Blazek.
\newblock Contactless mapping of rhythmical phenomena in tissue perfusion using ppgi.
\newblock In \emph{Medical Imaging 2002: Physiology and Function from Multidimensional Images}, pages 110--117. International Society for Optics and Photonics, 2002.

\bibitem[Jiang et~al.(2022{\natexlab{a}})Jiang, Murdock, and Ithapu]{jiang2022egocentric}
Hao Jiang, Calvin Murdock, and Vamsi~Krishna Ithapu.
\newblock Egocentric deep multi-channel audio-visual active speaker localization.
\newblock In \emph{Proceedings of the IEEE/CVF Conference on Computer Vision and Pattern Recognition}, pages 10544--10552, 2022{\natexlab{a}}.

\bibitem[Jiang et~al.(2022{\natexlab{b}})Jiang, Streli, Qiu, Fender, Laich, Snape, and Holz]{jiang2022avatarposer}
Jiaxi Jiang, Paul Streli, Huajian Qiu, Andreas Fender, Larissa Laich, Patrick Snape, and Christian Holz.
\newblock Avatarposer: Articulated full-body pose tracking from sparse motion sensing.
\newblock In \emph{European conference on computer vision}, pages 443--460. Springer, 2022{\natexlab{b}}.

\bibitem[Jiang et~al.(2023)Jiang, Streli, Meier, Fender, and Holz]{jiang2023egoposer}
Jiaxi Jiang, Paul Streli, Manuel Meier, Andreas Fender, and Christian Holz.
\newblock Egoposer: Robust real-time ego-body pose estimation in large scenes.
\newblock \emph{arXiv preprint arXiv:2308.06493}, 3\penalty0 (7), 2023.

\bibitem[Joshi et~al.(2024)Joshi, Agaian, and Cho]{joshi2024factorizephys}
Jitesh Joshi, Sos~S Agaian, and Youngjun Cho.
\newblock Factorizephys: Matrix factorization for multidimensional attention in remote physiological sensing.
\newblock \emph{arXiv preprint arXiv:2411.01542}, 2024.

\bibitem[Kendall et~al.(2015)Kendall, Grimes, and Cipolla]{kendall2015posenet}
Alex Kendall, Matthew Grimes, and Roberto Cipolla.
\newblock Posenet: A convolutional network for real-time 6-dof camera relocalization.
\newblock In \emph{Proceedings of the IEEE international conference on computer vision}, pages 2938--2946, 2015.

\bibitem[Kleiger et~al.(2005)Kleiger, Stein, and Bigger~Jr]{kleiger2005heart}
Robert~E Kleiger, Phyllis~K Stein, and J~Thomas Bigger~Jr.
\newblock Heart rate variability: measurement and clinical utility.
\newblock \emph{Annals of Noninvasive Electrocardiology}, 10\penalty0 (1):\penalty0 88--101, 2005.

\bibitem[Kondratyuk et~al.(2021)Kondratyuk, Yuan, Li, Zhang, Tan, Brown, and Gong]{kondratyuk2021movinets}
Dan Kondratyuk, Liangzhe Yuan, Yandong Li, Li Zhang, Mingxing Tan, Matthew Brown, and Boqing Gong.
\newblock Movinets: Mobile video networks for efficient video recognition.
\newblock In \emph{Proceedings of the IEEE/CVF conference on computer vision and pattern recognition}, pages 16020--16030, 2021.

\bibitem[Kovacevic et~al.(2024)Kovacevic, Holz, Gross, and Wampfler]{kovacevic2024multimodal}
Nikola Kovacevic, Christian Holz, Markus Gross, and Rafael Wampfler.
\newblock On multimodal emotion recognition for human-chatbot interaction in the wild.
\newblock In \emph{Proceedings of the 26th International Conference on Multimodal Interaction}, pages 12--21, 2024.

\bibitem[Leap(2024)]{magic_leap}
Magic Leap.
\newblock Magic leap 2.
\newblock \url{https://www.magicleap.com/magic-leap-2}, 2024.
\newblock Accessed: 2024.11.13.

\bibitem[Li et~al.(2018)Li, Alikhani, Shi, Seppanen, Junttila, Majamaa-Voltti, Tulppo, and Zhao]{li2018obf}
Xiaobai Li, Iman Alikhani, Jingang Shi, Tapio Seppanen, Juhani Junttila, Kirsi Majamaa-Voltti, Mikko Tulppo, and Guoying Zhao.
\newblock The obf database: A large face video database for remote physiological signal measurement and atrial fibrillation detection.
\newblock In \emph{2018 13th IEEE international conference on automatic face \& gesture recognition (FG 2018)}, pages 242--249. IEEE, 2018.

\bibitem[Liu et~al.(2020)Liu, Fromm, Patel, and McDuff]{liu2020multi}
Xin Liu, Josh Fromm, Shwetak Patel, and Daniel McDuff.
\newblock Multi-task temporal shift attention networks for on-device contactless vitals measurement.
\newblock \emph{Advances in Neural Information Processing Systems}, 33:\penalty0 19400--19411, 2020.

\bibitem[Liu et~al.(2024)Liu, Narayanswamy, Paruchuri, Zhang, Tang, Zhang, Sengupta, Patel, Wang, and McDuff]{liu2024rppg}
Xin Liu, Girish Narayanswamy, Akshay Paruchuri, Xiaoyu Zhang, Jiankai Tang, Yuzhe Zhang, Roni Sengupta, Shwetak Patel, Yuntao Wang, and Daniel McDuff.
\newblock rppg-toolbox: Deep remote ppg toolbox.
\newblock \emph{Advances in Neural Information Processing Systems}, 36, 2024.

\bibitem[Lugaresi et~al.(2019)Lugaresi, Tang, Nash, McClanahan, Uboweja, Hays, Zhang, Chang, Yong, Lee, et~al.]{lugaresi2019mediapipe}
Camillo Lugaresi, Jiuqiang Tang, Hadon Nash, Chris McClanahan, Esha Uboweja, Michael Hays, Fan Zhang, Chuo-Ling Chang, Ming Yong, Juhyun Lee, et~al.
\newblock Mediapipe: A framework for perceiving and processing reality.
\newblock In \emph{Third workshop on computer vision for AR/VR at IEEE computer vision and pattern recognition (CVPR)}, 2019.

\bibitem[Luong and Holz(2022)]{luong2022characterizing}
Tiffany Luong and Christian Holz.
\newblock Characterizing physiological responses to fear, frustration, and insight in virtual reality.
\newblock \emph{IEEE Transactions on Visualization and Computer Graphics}, 28\penalty0 (11):\penalty0 3917--3927, 2022.

\bibitem[Lv et~al.(2024)Lv, Charron, Moulon, Gamino, Peng, Sweeney, Miller, Tang, Meissner, Dong, et~al.]{lv2024aria}
Zhaoyang Lv, Nicholas Charron, Pierre Moulon, Alexander Gamino, Cheng Peng, Chris Sweeney, Edward Miller, Huixuan Tang, Jeff Meissner, Jing Dong, et~al.
\newblock Aria everyday activities dataset.
\newblock \emph{arXiv preprint arXiv:2402.13349}, 2024.

\bibitem[Ma et~al.(2024)Ma, Ye, Hong, Guzov, Jiang, Postyeni, Pesqueira, Gamino, Baiyya, Kim, et~al.]{ma2024nymeria}
Lingni Ma, Yuting Ye, Fangzhou Hong, Vladimir Guzov, Yifeng Jiang, Rowan Postyeni, Luis Pesqueira, Alexander Gamino, Vijay Baiyya, Hyo~Jin Kim, et~al.
\newblock Nymeria: A massive collection of multimodal egocentric daily motion in the wild.
\newblock \emph{arXiv preprint arXiv:2406.09905}, 2024.

\bibitem[Ma et~al.(2016)Ma, Fan, and Kitani]{ma2016going}
Minghuang Ma, Haoqi Fan, and Kris~M Kitani.
\newblock Going deeper into first-person activity recognition.
\newblock In \emph{Proceedings of the IEEE Conference on Computer Vision and Pattern Recognition}, pages 1894--1903, 2016.

\bibitem[MacPherson et~al.(2009)MacPherson, Collins, and Obhi]{macpherson2009importance}
Alan~C MacPherson, Dave Collins, and Sukhvinder~S Obhi.
\newblock The importance of temporal structure and rhythm for the optimum performance of motor skills: A new focus for practitioners of sport psychology.
\newblock \emph{Journal of Applied Sport Psychology}, 21\penalty0 (S1):\penalty0 S48--S61, 2009.

\bibitem[Mar{\'\i}n-Morales et~al.(2020)Mar{\'\i}n-Morales, Llinares, Guixeres, and Alca{\~n}iz]{marin2020emotion}
Javier Mar{\'\i}n-Morales, Carmen Llinares, Jaime Guixeres, and Mariano Alca{\~n}iz.
\newblock Emotion recognition in immersive virtual reality: From statistics to affective computing.
\newblock \emph{Sensors}, 20\penalty0 (18):\penalty0 5163, 2020.

\bibitem[McDuff et~al.(2022)McDuff, Wander, Liu, Hill, Hernandez, Lester, and Baltrusaitis]{mcduff2022scamps}
Daniel McDuff, Miah Wander, Xin Liu, Brian Hill, Javier Hernandez, Jonathan Lester, and Tadas Baltrusaitis.
\newblock Scamps: Synthetics for camera measurement of physiological signals.
\newblock \emph{Advances in Neural Information Processing Systems}, 35:\penalty0 3744--3757, 2022.

\bibitem[Meier et~al.(2025)Meier, Demirel, and Holz]{meier2024wildppg}
Manuel Meier, Berken~Utku Demirel, and Christian Holz.
\newblock Wildppg: a real-world ppg dataset of long continuous recordings.
\newblock In \emph{Proceedings of the 38th International Conference on Neural Information Processing Systems}, Red Hook, NY, USA, 2025. Curran Associates Inc.

\bibitem[Meta(2024)]{meta_quest}
Meta.
\newblock Meta quest.
\newblock \url{https://www.meta.com/quest/}, 2024.
\newblock Accessed: 2024.11.13.

\bibitem[Microsoft(2024)]{microsoft_hololens}
Microsoft.
\newblock Microsoft hololens.
\newblock \url{https://learn.microsoft.com/en-us/hololens/}, 2024.
\newblock Accessed: 2024.11.13.

\bibitem[Min and Corso(2021)]{min2021integrating}
Kyle Min and Jason~J Corso.
\newblock Integrating human gaze into attention for egocentric activity recognition.
\newblock In \emph{Proceedings of the IEEE/CVF Winter Conference on Applications of Computer Vision}, pages 1069--1078, 2021.

\bibitem[Miranda-Correa et~al.(2018)Miranda-Correa, Abadi, Sebe, and Patras]{miranda2018amigos}
Juan~Abdon Miranda-Correa, Mojtaba~Khomami Abadi, Nicu Sebe, and Ioannis Patras.
\newblock Amigos: A dataset for affect, personality and mood research on individuals and groups.
\newblock \emph{IEEE transactions on affective computing}, 12\penalty0 (2):\penalty0 479--493, 2018.

\bibitem[Moebus et~al.(2024{\natexlab{a}})Moebus, Gashi, Hilty, Oldrati, and Holz]{moebus2024meaningful}
Max Moebus, Shkurta Gashi, Marc Hilty, Pietro Oldrati, and Christian Holz.
\newblock Meaningful digital biomarkers derived from wearable sensors to predict daily fatigue in multiple sclerosis patients and healthy controls.
\newblock \emph{Iscience}, 27\penalty0 (2), 2024{\natexlab{a}}.

\bibitem[Moebus et~al.(2024{\natexlab{b}})Moebus, Hauptmann, Kopp, Demirel, Braun, and Holz]{moebus2024nightbeat}
Max Moebus, Lars Hauptmann, Nicolas Kopp, Berken Demirel, Bj{\"o}rn Braun, and Christian Holz.
\newblock Nightbeat: Heart rate estimation from a wrist-worn accelerometer during sleep.
\newblock \emph{IEEE Journal of Biomedical and Health Informatics}, 2024{\natexlab{b}}.

\bibitem[Mukhopadhyay(2014)]{mukhopadhyay2014wearable}
Subhas~Chandra Mukhopadhyay.
\newblock Wearable sensors for human activity monitoring: A review.
\newblock \emph{IEEE sensors journal}, 15\penalty0 (3):\penalty0 1321--1330, 2014.

\bibitem[Niu et~al.(2019{\natexlab{a}})Niu, Han, Shan, and Chen]{niu2019vipl}
Xuesong Niu, Hu Han, Shiguang Shan, and Xilin Chen.
\newblock Vipl-hr: A multi-modal database for pulse estimation from less-constrained face video.
\newblock In \emph{Computer Vision--ACCV 2018: 14th Asian Conference on Computer Vision, Perth, Australia, December 2--6, 2018, Revised Selected Papers, Part V 14}, pages 562--576. Springer, 2019{\natexlab{a}}.

\bibitem[Niu et~al.(2019{\natexlab{b}})Niu, Zhao, Han, Das, Dantcheva, Shan, and Chen]{niu2019robust}
Xuesong Niu, Xingyuan Zhao, Hu Han, Abhijit Das, Antitza Dantcheva, Shiguang Shan, and Xilin Chen.
\newblock Robust remote heart rate estimation from face utilizing spatial-temporal attention.
\newblock In \emph{2019 14th IEEE international conference on automatic face \& gesture recognition (FG 2019)}, pages 1--8. IEEE, 2019{\natexlab{b}}.

\bibitem[Picard(2000)]{picard2000affective}
Rosalind~W Picard.
\newblock \emph{Affective computing}.
\newblock MIT press, 2000.

\bibitem[Poh et~al.(2010)Poh, McDuff, and Picard]{poh2010non}
Ming-Zher Poh, Daniel McDuff, and Rosalind~W Picard.
\newblock Non-contact, automated cardiac pulse measurements using video imaging and blind source separation.
\newblock \emph{Optics express}, 18\penalty0 (10):\penalty0 10762--10774, 2010.

\bibitem[Rodin et~al.(2021)Rodin, Furnari, Mavroeidis, and Farinella]{rodin2021predicting}
Ivan Rodin, Antonino Furnari, Dimitrios Mavroeidis, and Giovanni~Maria Farinella.
\newblock Predicting the future from first person (egocentric) vision: A survey.
\newblock \emph{Computer Vision and Image Understanding}, 211:\penalty0 103252, 2021.

\bibitem[Rosinol et~al.(2023)Rosinol, Leonard, and Carlone]{rosinol2023nerf}
Antoni Rosinol, John~J Leonard, and Luca Carlone.
\newblock Nerf-slam: Real-time dense monocular slam with neural radiance fields.
\newblock In \emph{2023 IEEE/RSJ International Conference on Intelligent Robots and Systems (IROS)}, pages 3437--3444. IEEE, 2023.

\bibitem[Ryoo et~al.(2015)Ryoo, Fuchs, Xia, Aggarwal, and Matthies]{ryoo2015robot}
Michael~S Ryoo, Thomas~J Fuchs, Lu Xia, Jake~K Aggarwal, and Larry Matthies.
\newblock Robot-centric activity prediction from first-person videos: What will they do to me?
\newblock In \emph{Proceedings of the tenth annual ACM/IEEE international conference on human-robot interaction}, pages 295--302, 2015.

\bibitem[Sabour et~al.(2021)Sabour, Benezeth, De~Oliveira, Chappe, and Yang]{sabour2021ubfc}
Rita~Meziati Sabour, Yannick Benezeth, Pierre De~Oliveira, Julien Chappe, and Fan Yang.
\newblock Ubfc-phys: A multimodal database for psychophysiological studies of social stress.
\newblock \emph{IEEE Transactions on Affective Computing}, 2021.

\bibitem[Sattler et~al.(2011)Sattler, Leibe, and Kobbelt]{sattler2011fast}
Torsten Sattler, Bastian Leibe, and Leif Kobbelt.
\newblock Fast image-based localization using direct 2d-to-3d matching.
\newblock In \emph{2011 International Conference on Computer Vision}, pages 667--674. IEEE, 2011.

\bibitem[Shaffer and Ginsberg(2017)]{shaffer2017overview}
Fred Shaffer and Jay~P Ginsberg.
\newblock An overview of heart rate variability metrics and norms.
\newblock \emph{Frontiers in public health}, 5:\penalty0 258, 2017.

\bibitem[Shan et~al.(2020)Shan, Geng, Shu, and Fouhey]{shan2020understanding}
Dandan Shan, Jiaqi Geng, Michelle Shu, and David~F Fouhey.
\newblock Understanding human hands in contact at internet scale.
\newblock In \emph{Proceedings of the IEEE/CVF conference on computer vision and pattern recognition}, pages 9869--9878, 2020.

\bibitem[Shavit et~al.(2021)Shavit, Ferens, and Keller]{shavit2021learning}
Yoli Shavit, Ron Ferens, and Yosi Keller.
\newblock Learning multi-scene absolute pose regression with transformers.
\newblock In \emph{Proceedings of the IEEE/CVF International Conference on Computer Vision}, pages 2733--2742, 2021.

\bibitem[Shiratori et~al.(2011)Shiratori, Park, Sigal, Sheikh, and Hodgins]{shiratori2011motion}
Takaaki Shiratori, Hyun~Soo Park, Leonid Sigal, Yaser Sheikh, and Jessica~K Hodgins.
\newblock Motion capture from body-mounted cameras.
\newblock In \emph{ACM SIGGRAPH 2011 papers}, pages 1--10, 2011.

\bibitem[Shuggi et~al.(2019)Shuggi, Oh, Wu, Ayoub, Moreno, Shaw, Shewokis, and Gentili]{shuggi2019motor}
Isabelle~M Shuggi, Hyuk Oh, Helena Wu, Maria~J Ayoub, Arianna Moreno, Emma~P Shaw, Patricia~A Shewokis, and Rodolphe~J Gentili.
\newblock Motor performance, mental workload and self-efficacy dynamics during learning of reaching movements throughout multiple practice sessions.
\newblock \emph{Neuroscience}, 423:\penalty0 232--248, 2019.

\bibitem[Soleymani et~al.(2011)Soleymani, Lichtenauer, Pun, and Pantic]{soleymani2011multimodal}
Mohammad Soleymani, Jeroen Lichtenauer, Thierry Pun, and Maja Pantic.
\newblock A multimodal database for affect recognition and implicit tagging.
\newblock \emph{IEEE transactions on affective computing}, 3\penalty0 (1):\penalty0 42--55, 2011.

\bibitem[Speth et~al.(2023)Speth, Vance, Flynn, and Czajka]{speth2023non}
Jeremy Speth, Nathan Vance, Patrick Flynn, and Adam Czajka.
\newblock Non-contrastive unsupervised learning of physiological signals from video.
\newblock In \emph{Proceedings of the IEEE/CVF Conference on Computer Vision and Pattern Recognition}, pages 14464--14474, 2023.

\bibitem[Stricker et~al.(2014)Stricker, Müller, and Gross]{stricker2014pure}
Ronny Stricker, Steffen Müller, and Horst-Michael Gross.
\newblock Non-contact video-based pulse rate measurement on a mobile service robot.
\newblock In \emph{The 23rd IEEE International Symposium on Robot and Human Interactive Communication}, pages 1056--1062, 2014.

\bibitem[Subramanian et~al.(2016)Subramanian, Wache, Abadi, Vieriu, Winkler, and Sebe]{subramanian2016ascertain}
Ramanathan Subramanian, Julia Wache, Mojtaba~Khomami Abadi, Radu~L Vieriu, Stefan Winkler, and Nicu Sebe.
\newblock Ascertain: Emotion and personality recognition using commercial sensors.
\newblock \emph{IEEE Transactions on Affective Computing}, 9\penalty0 (2):\penalty0 147--160, 2016.

\bibitem[Sun and Li(2024)]{sun2024contrast}
Zhaodong Sun and Xiaobai Li.
\newblock Contrast-phys+: Unsupervised and weakly-supervised video-based remote physiological measurement via spatiotemporal contrast.
\newblock \emph{IEEE Transactions on Pattern Analysis and Machine Intelligence}, 2024.

\bibitem[Tang et~al.(2023)Tang, Chen, Wang, Shi, Patel, McDuff, and Liu]{tang2023mmpd}
Jiankai Tang, Kequan Chen, Yuntao Wang, Yuanchun Shi, Shwetak Patel, Daniel McDuff, and Xin Liu.
\newblock Mmpd: multi-domain mobile video physiology dataset.
\newblock In \emph{2023 45th Annual International Conference of the IEEE Engineering in Medicine \& Biology Society (EMBC)}, pages 1--5. IEEE, 2023.

\bibitem[Tyng et~al.(2017)Tyng, Amin, Saad, and Malik]{tyng2017influences}
Chai~M Tyng, Hafeez~U Amin, Mohamad~NM Saad, and Aamir~S Malik.
\newblock The influences of emotion on learning and memory.
\newblock \emph{Frontiers in psychology}, 8:\penalty0 235933, 2017.

\bibitem[Verkruysse et~al.(2008)Verkruysse, Svaasand, and Nelson]{verkruysse2008remote}
Wim Verkruysse, Lars~O Svaasand, and J~Stuart Nelson.
\newblock Remote plethysmographic imaging using ambient light.
\newblock \emph{Optics express}, 16\penalty0 (26):\penalty0 21434--21445, 2008.

\bibitem[Wang et~al.(2016)Wang, Den~Brinker, Stuijk, and De~Haan]{wang2016algorithmic}
Wenjin Wang, Albertus~C Den~Brinker, Sander Stuijk, and Gerard De~Haan.
\newblock Algorithmic principles of remote ppg.
\newblock \emph{IEEE Transactions on Biomedical Engineering}, 64\penalty0 (7):\penalty0 1479--1491, 2016.

\bibitem[Woo et~al.(2018)Woo, Park, Lee, and Kweon]{woo2018cbam}
Sanghyun Woo, Jongchan Park, Joon-Young Lee, and In~So Kweon.
\newblock Cbam: Convolutional block attention module.
\newblock In \emph{Proceedings of the European conference on computer vision (ECCV)}, pages 3--19, 2018.

\bibitem[Wu et~al.(2022)Wu, Li, Mangalam, Fan, Xiong, Malik, and Feichtenhofer]{wu2022memvit}
Chao-Yuan Wu, Yanghao Li, Karttikeya Mangalam, Haoqi Fan, Bo Xiong, Jitendra Malik, and Christoph Feichtenhofer.
\newblock Memvit: Memory-augmented multiscale vision transformer for efficient long-term video recognition.
\newblock In \emph{Proceedings of the IEEE/CVF Conference on Computer Vision and Pattern Recognition}, pages 13587--13597, 2022.

\bibitem[Yamada et~al.(2012)Yamada, Sugano, Okabe, Sato, Sugimoto, and Hiraki]{yamada2012attention}
Kentaro Yamada, Yusuke Sugano, Takahiro Okabe, Yoichi Sato, Akihiro Sugimoto, and Kazuo Hiraki.
\newblock Attention prediction in egocentric video using motion and visual saliency.
\newblock In \emph{Advances in Image and Video Technology: 5th Pacific Rim Symposium, PSIVT 2011, Gwangju, South Korea, November 20-23, 2011, Proceedings, Part I 5}, pages 277--288. Springer, 2012.

\bibitem[Yan et~al.(2022)Yan, Xiong, Arnab, Lu, Zhang, Sun, and Schmid]{yan2022multiview}
Shen Yan, Xuehan Xiong, Anurag Arnab, Zhichao Lu, Mi Zhang, Chen Sun, and Cordelia Schmid.
\newblock Multiview transformers for video recognition.
\newblock In \emph{Proceedings of the IEEE/CVF conference on computer vision and pattern recognition}, pages 3333--3343, 2022.

\bibitem[Yan et~al.(2024)Yan, Zhong, Zhang, Shu, Xu, and Kang]{yan2024physmamba}
Zhixin Yan, Yan Zhong, Wenjun Zhang, Lin Shu, Hongbin Xu, and Wenxiong Kang.
\newblock Physmamba: Leveraging dual-stream cross-attention ssd for remote physiological measurement.
\newblock \emph{arXiv preprint arXiv:2408.01077}, 2024.

\bibitem[Yao et~al.(2019)Yao, Xu, Choi, Crandall, Atkins, and Dariush]{yao2019egocentric}
Yu Yao, Mingze Xu, Chiho Choi, David~J Crandall, Ella~M Atkins, and Behzad Dariush.
\newblock Egocentric vision-based future vehicle localization for intelligent driving assistance systems.
\newblock In \emph{2019 International Conference on Robotics and Automation (ICRA)}, pages 9711--9717. IEEE, 2019.

\bibitem[Yu et~al.(2019)Yu, Li, and Zhao]{yu2019physnet}
Zitong Yu, Xiaobai Li, and Guoying Zhao.
\newblock Remote photoplethysmograph signal measurement from facial videos using spatio-temporal networks.
\newblock \emph{arXiv preprint arXiv:1905.02419}, 2019.

\bibitem[Yu et~al.(2022)Yu, Shen, Shi, Zhao, Torr, and Zhao]{yu2022physformer}
Zitong Yu, Yuming Shen, Jingang Shi, Hengshuang Zhao, Philip~HS Torr, and Guoying Zhao.
\newblock Physformer: Facial video-based physiological measurement with temporal difference transformer.
\newblock In \emph{Proceedings of the IEEE/CVF conference on computer vision and pattern recognition}, pages 4186--4196, 2022.

\bibitem[Yue et~al.(2023)Yue, Shi, and Ding]{yue2023facial}
Zijie Yue, Miaojing Shi, and Shuai Ding.
\newblock Facial video-based remote physiological measurement via self-supervised learning.
\newblock \emph{IEEE Transactions on Pattern Analysis and Machine Intelligence}, 45\penalty0 (11):\penalty0 13844--13859, 2023.

\bibitem[Zhang et~al.(2016)Zhang, Girard, Wu, Zhang, Liu, Ciftci, Canavan, Reale, Horowitz, Yang, et~al.]{zhang2016multimodal}
Zheng Zhang, Jeff~M Girard, Yue Wu, Xing Zhang, Peng Liu, Umur Ciftci, Shaun Canavan, Michael Reale, Andy Horowitz, Huiyuan Yang, et~al.
\newblock Multimodal spontaneous emotion corpus for human behavior analysis.
\newblock In \emph{Proceedings of the IEEE conference on computer vision and pattern recognition}, pages 3438--3446, 2016.

\bibitem[Zhao et~al.(2023)Zhao, Misra, Kr{\"a}henb{\"u}hl, and Girdhar]{zhao2023learning}
Yue Zhao, Ishan Misra, Philipp Kr{\"a}henb{\"u}hl, and Rohit Girdhar.
\newblock Learning video representations from large language models.
\newblock In \emph{Proceedings of the IEEE/CVF Conference on Computer Vision and Pattern Recognition}, pages 6586--6597, 2023.

\bibitem[Zhu et~al.(2023)Zhu, Xiao, Alvarado, Babaei, Hu, El-Mohri, Culatana, Sumbaly, and Yan]{zhu2023egoobjects}
Chenchen Zhu, Fanyi Xiao, Andr{\'e}s Alvarado, Yasmine Babaei, Jiabo Hu, Hichem El-Mohri, Sean Culatana, Roshan Sumbaly, and Zhicheng Yan.
\newblock Egoobjects: A large-scale egocentric dataset for fine-grained object understanding.
\newblock In \emph{Proceedings of the IEEE/CVF International Conference on Computer Vision}, pages 20110--20120, 2023.

\bibitem[Zou et~al.(2024{\natexlab{a}})Zou, Guo, Chen, and Ma]{zou2024rhythmformer}
Bochao Zou, Zizheng Guo, Jiansheng Chen, and Huimin Ma.
\newblock Rhythmformer: Extracting rppg signals based on hierarchical temporal periodic transformer.
\newblock \emph{arXiv preprint arXiv:2402.12788}, 2024{\natexlab{a}}.

\bibitem[Zou et~al.(2024{\natexlab{b}})Zou, Guo, Hu, and Ma]{zou2024rhythmmamba}
Bochao Zou, Zizheng Guo, Xiaocheng Hu, and Huimin Ma.
\newblock Rhythmmamba: Fast remote physiological measurement with arbitrary length videos.
\newblock \emph{arXiv preprint arXiv:2404.06483}, 2024{\natexlab{b}}.

\end{thebibliography}
